\documentclass[10pt]{article} 
\usepackage[preprint]{tmlr}

\usepackage{amsmath,amsfonts,bm}

\def\eqref#1{equation~\ref{#1}}

\def\1{\bm{1}}

\DeclareMathAlphabet{\mathsfit}{\encodingdefault}{\sfdefault}{m}{sl}
\SetMathAlphabet{\mathsfit}{bold}{\encodingdefault}{\sfdefault}{bx}{n}

\usepackage{hyperref}
\usepackage{url}
\usepackage{xcolor}
\usepackage{titletoc}

\title{Multimodal Concept Bottleneck Models}

\author{\name Tongqing Shi \email toshi@ucsd.edu \\
      \addr Halıcıoğlu Data Science Institute \\
      University of California, San Diego
      \AND
      \name Ge Yan \email geyan@ucsd.edu \\
      \addr Department of Computer Science and Engineering \\
      University of California, San Diego
      \AND
      \name Tuomas Oikarinen \email toikarinen@ucsd.edu\\
      \addr Department of Computer Science and Engineering \\
      University of California, San Diego
      \AND
      \name Tsui-Wei Weng \email lweng@ucsd.edu \\
      \addr Halıcıoğlu Data Science Institute \\
      University of California, San Diego
      }

\def\month{10}  
\def\year{2026} 
\def\openreview{\url{https://openreview.net/forum?id=wBagHznSw7}} 

\begin{document}

\maketitle

\begin{abstract}
Concept Bottleneck Models (CBMs) enhance the interpretability of deep neural networks by aligning the features extracted from images with human-interpretable concepts. However, conventional CBMs typically use a linear classifier that maps concept activations to a fixed set of predefined classes, limiting their flexibility in open-vocabulary settings. In this paper, we propose Multimodal Concept Bottleneck Model (MM-CBM) to address these issues and extend CBMs into CLIP structure. MM-CBM utilizes dual Concept Bottleneck Layers (CBLs) to align both the image and text embeddings into interpretable features. This allows us to perform new vision tasks like zero-shot classification or image retrieval in an interpretable way. Compared to existing methods, MM-CBM achieves 23.28\% accuracy improvement on average across four standard benchmarks. Our method maintains competitive performance compared with black-box CLIP model while offering interpretability. Under classification tasks, MM-CBM's accuracy is on average 3.94\% lower than CLIP in the target-dataset adaptation setting and 9.84\% in the zero-shot transfer setting. Under image-text retrieval tasks, it achieves an average gap of 2.48\% for image retrieval and 9.22\% for text retrieval.

\end{abstract}

\section{Introduction}
\label{sec:introduction}

\begin{figure}[h]
  \centering
  \includegraphics[width=\linewidth]{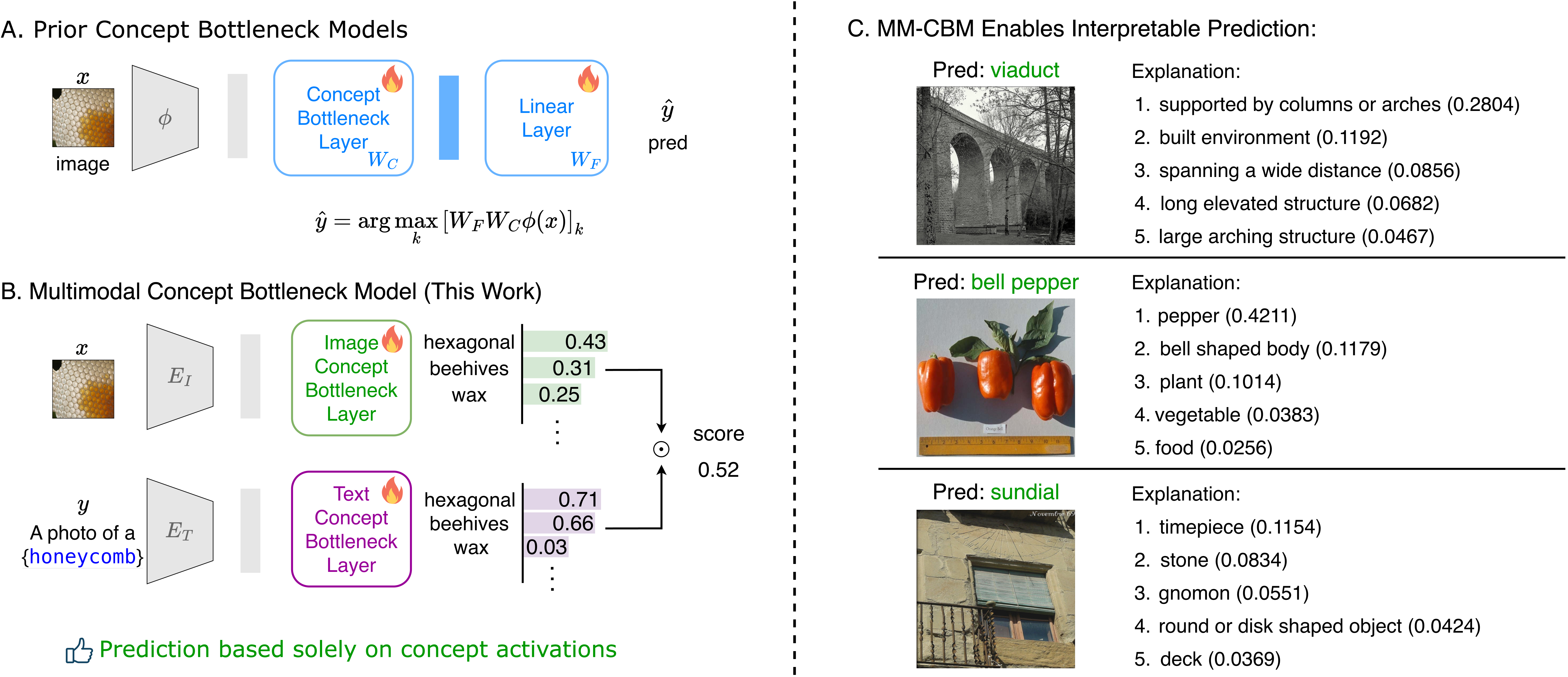}
  \caption{
    \textbf{A.} Previous workflow of concept bottleneck models.
    \textbf{B.} Our MM-CBM makes predictions based solely on concept responses.
    \textbf{C.} Visualization of explanations on several samples.
  }
  \label{fig: Comparison of model}
\vspace{-0.3cm}

\end{figure}

The opacity and lack of interpretability of deep learning models hinder their deployment in real-world applications. To mitigate this, numerous post-hoc neuron-level explanation methods have been proposed, aiming to understand the semantics of individual neurons~\citep{goh2021multimodal, oikarinen2024linear, bau2017network, hernandez2021natural, khan2023concept, shaham2024multimodal, bills2023language}. However, these methods often struggle with polysemanticity, where a single neuron encodes multiple, potentially unrelated concepts, limiting their reliability.

As an alternative, researchers have developed intrinsically interpretable models such as Concept Bottleneck Models (CBMs)~\citep{koh2020concept, oikarinen2023label, yan2023learning, srivastavavlg, yang2023language}. CBMs introduce a human-interpretable concept bottleneck layer (CBL) before the classifier, ensuring that predictions are grounded in semantically meaningful concepts. Despite their advantages, existing CBMs rely on a linear classifier placed after the CBL, which introduces two limitations:
(1) \textbf{Restriction to predefined class labels:} Conventional CBMs can only predict among a fixed set of categories and cannot support natural language queries.  
(2) \textbf{Limited reliance on learned concepts:} Recent work~\citep{yan2023learning, srivastavavlg} shows that the downstream linear layer may circumvent concept activations, allowing the model to achieve high accuracy even if the CBL weights are randomly initialized.

On the other hand, powerful vision-language models such as CLIP overcome the limitation of fixed class label sets by performing open-vocabulary recognition through image–text embedding similarity. However, the reasoning behind CLIP’s decisions remains opaque: the model does not reveal which human-understandable concepts contribute to a high similarity score. This makes it difficult to interpret why a prediction is made or diagnose model failures. In other words, CBMs provide interpretability but lack flexibility, while CLIP is flexible but lacks interpretability. A natural question arises:
\textit{Can we design an architecture that supports open-vocabulary predictions while maintaining full transparency in the decision-making process?}

To address this, we propose a new framework called \textbf{Multimodal Concept Bottleneck Model (MM-CBM)}. Unlike existing CBMs that rely on a single CBL, MM-CBM introduces \textbf{dual CBLs}, one for the image inputs and the other for text input, ensuring that the same conceptual semantics are aligned across both modalities. The text encoder, together with the text CBL, converts arbitrary textual inputs into concept responses; at the same time, the image encoder, together with the image CBL, maps visual inputs into the same concept space as shown in Fig.~\ref{fig: Comparison of model}B. During inference, predictions are made by computing the similarity between concept responses from both modalities. Additionally, by leveraging CLIP's rich semantic knowledge, MM-CBM maintains competitive performance relative to black-box CLIP while providing concept-level interpretability.

\noindent \textbf{Our key contributions are summarized as follows:}
\begin{itemize}
    \item We propose the first \textbf{dual concept bottleneck layers} CBM framework that incorporates concept bottleneck layers across both visual and textual modalities. This design enables complex multimodal tasks such as zero-shot generalization, image retrieval, and open-vocabulary recognition, while preserving concept-level interpretability. 
    
    \item We establish a \textbf{fully inspectable reasoning pipeline} in which all final predictions are derived solely from interpretable concept responses. By integrating the concept bottleneck mechanism into a vision-language framework, our method enables CLIP-style image--text matching through an explicit and inspectable concept space, providing human-understandable concept-level explanations while maintaining open-vocabulary flexibility.
    
    \item Extensive experiments demonstrate that our framework significantly outperforms existing CBM methods under the ANEC-5 metric (accuracy under $\mathrm{NEC}=5$, a metric proposed by~\cite{srivastavavlg} for fair comparison of CBMs), reaching 71.08\% across four benchmarks and outperforming the strongest baseline, VLG-CBM, by 2.48\%. Furthermore, our model maintains competitive performance with CLIP across supervised adaptation, zero-shot transfer, and image-text retrieval settings while offering fully concept-mediated prediction.
\end{itemize}

\section{Related work}
\label{sec:relatedwork}

\noindent \textbf{Post-hoc neuron-level explanations.} 
Recent advances in representation-based post-hoc explanation methods have provided new insights into understanding neural network behavior at a global level. \cite{bau2017network} aligned neuron activations with human-labeled image regions using manually annotated datasets, thereby assigning semantic concepts to individual neurons. \cite{kalibhat2023identifying, hernandez2021natural} identified highly activated image regions through predefined submodules or image captioning models to generate neuron-level explanations.
CLIP-Dissect~\citep{oikarinenclip} associates neuron activations with predefined textual concepts, while Linear Explanations~\citep{oikarinen2024linear} captures neuronal polysemanticity by modeling individual neurons as linear combinations of concepts. Describe-and-Dissect~\citep{bai2025interpret} generates natural-language descriptions without a predefined concept vocabulary. The inherent polysemanticity of neurons in modern neural networks forms the foundation upon which we build our CBL: by linearly combining neuron activations, we are able to synthesize clear, human-aligned concepts.

\noindent \textbf{Concept bottleneck models (CBMs).} 
CBMs~\citep{koh2020concept} aim to build intrinsically interpretable models by aligning intermediate representations with human-understandable concepts. A typical CBM architecture consists of two components: a \textit{concept predictor} and a \textit{class predictor}. Given an input $x \in \mathcal{X}$ and a feature extractor $\phi$, the model first maps extracted features $\phi(x)$ to concept activations $\hat{a}(x) = W_C \phi(x) \in \mathbb{R}^{|C|}$ via an embedding matrix $W_C$, where $C$ denotes the set of candidate concepts. Each coordinate of $\hat{a}(x)$ corresponds to a specific interpretable concept. The final prediction is then obtained by applying a linear classifier parameterized by $W_F$ on top of the concept space. This yields a modified prediction formulation of the form:

\begin{equation}
    \hat y=\arg\max_k \left[W_FW_C\phi(x)\right]_k.
\end{equation}

This architecture introduces an explicit intermediate concept representation for class prediction, allowing concept activations to be inspected and modified through interventions. Recent works~\citep{yuksekgonulpost, oikarinen2023label, yang2023language, yan2023learning, srivastavavlg, hu2025semi, yu2025language, liu2025hybrid, zhang2025attribute} have also explored more flexible training paradigms to improve the generalization and applicability of CBMs, such as leveraging weak supervision, soft labels, or multimodal signals. Concept bottlenecks have further been extended to large language models like CB-LLM~\citep{sun2024concept} and image generation through CB-AE~\citep{kulkarni2025interpretable}; our work focuses on the visual classification task. However, as pointed out by~\cite{yan2023learning, srivastavavlg}, when the dimensionality of the concept layer is sufficiently large, even randomly embedded features can suffice for a linear classifier to approximate the original prediction. That is, given a sufficiently high-dimensional projection $W_C$, even a randomly initialized one, it is possible to analytically recover a classifier $W_F$ such that $W_FW_C\phi(x)\approx W \phi(x)$, where $W$ is the original classifier, as shown in Fig.~\ref{fig: Comparison of model}A. This undermines the faithfulness and constraint role of the bottleneck layer, because accurate predictions may coexist with concept activations that do not faithfully represent their assigned meanings. In contrast, our architectural distinction is to replace fixed class-specific classifier weights with concept representations generated by a text CBL. Predictions are computed by matching image and text representations in the shared concept space, enabling new textual class queries without fitting a new classification head. A detailed comparison between our method and prior CBMs is provided in Table~\ref{table: comparison methods}.

\noindent \textbf{CLIP and its interpretability.}  
CLIP~\citep{radford2021learning} is a large-scale vision-language model trained on extensive image-text pairs using natural language supervision. It achieves strong zero-shot classification performance by encoding both images and text into a shared embedding space and computing similarity scores for prediction. Due to its strong generalization and semantic understanding capabilities, CLIP representations have been widely adopted in tasks such as semantic segmentation, object detection, visual question answering (VQA), and prompt generation for generative models. Numerous variants have been developed to enhance generalization~\citep{sun2023eva, zhai2023sigmoid} and computational efficiency~\citep{li2023scaling}.

Several efforts have also been made to interpret CLIP’s internal representations. \cite{goh2021multimodal} revealed the presence of multimodal polysemantic neurons within CLIP, showing that individual neurons can encode multiple abstract visual and textual concepts. \cite{bhalla2024interpreting} used dictionary learning to decompose CLIP representations into interpretable semantic components. \cite{menonvisual} proposed a training-free approach that leverages large language models (LLMs) to interpret CLIP’s predictions, thereby improving both transparency and performance. Our approach preserves CLIP's shared image--text representation structure while introducing predefined concepts as explicit dimensions, making concept-level interpretation intrinsic to the prediction process.

\begin{table}[t]
\centering
\caption{Comparative analysis of methods based on evaluation, flexibility, and interpretability. Here, $\checkmark$ denotes the method satisfies the requirement, $\triangle$ denotes the method partially satisfies the requirement, and $\times$ denotes the method does not satisfy the requirement. We compare with SOTA methods including LF-CBM~\citep{oikarinen2023label}, Labo~\citep{yang2023language}, LM4CV~\citep{yan2023learning} and VLG-CBM~\citep{srivastavavlg}.}

\resizebox{\textwidth}{!}{
\begin{tabular}{@{}l||cc|cc|cc@{}}
\toprule
& \multicolumn{2}{c|}{\textbf{Evaluation}} & \multicolumn{2}{c|}{\textbf{Flexibility}} & \multicolumn{2}{c}{\textbf{Interpretability}} \\ \cmidrule{2-7}
Method 
& Zero-shot & Sparse & Flexible &  Free text &  Free from  & Multimodal  \\ 
 & generalization & explanation & backbone & input & linear classifier & interpretability \\  \midrule
\textit{Baselines:} & & & & & \\ 
CLIP\citep{radford2021learning} & $\checkmark$ & $\times$ & $\checkmark$ & $\checkmark$ & $\checkmark$ & $\times$ \\
LF-CBM\citep{oikarinen2023label} & $\times$ & $\triangle$ & $\checkmark$ & $\times$ & $\times$ & $\triangle$ \\
LaBo\citep{yang2023language} & $\times$ & $\times$ & $\times$ & $\times$ & $\times$ & $\triangle$ \\
LM4CV\citep{yan2023learning} & $\times$ & $\times$ & $\times$ & $\times$ & $\times$ & $\triangle$ \\
VLG-CBM\citep{srivastavavlg} & $\times$ & $\checkmark$ & $\checkmark$ & $\times$ & $\times$ & $\triangle$ \\
\midrule 
\textit{This work:} & & & & & \\
MM-CBM & $\checkmark$ & $\checkmark$ & $\checkmark$ & $\checkmark$ & $\checkmark$ & $\checkmark$ \\ \bottomrule
\end{tabular}
}

\label{table: comparison methods}
\end{table}

\section{Proposed Method: MM-CBM}
\label{sec:method}

In this section, we introduce \textbf{Multimodal Concept Bottleneck Models (MM-CBM)}, a novel framework designed to improve the transparency and interpretability of multimodal reasoning by establishing dual Concept Bottleneck Layers (CBLs). Unlike traditional CBMs that rely on a final linear classification head, MM-CBM operates entirely within the concept space, thereby eliminating the dependence on the linear classifier and enabling fully inspectable post-bottleneck inference.

By incorporating text-based concept encodings, our approach supports a wider variety of natural language inputs, removing the limitation of fixed N-way classification and enabling open-vocabulary image-text matching and zero-shot generalization. MM-CBM is composed of three main stages:  
(1) \textbf{Collecting concept activation data},  
(2) \textbf{Training dual concept bottleneck layers}, and  
(3) \textbf{Performing inference at test time}, as described in Sections~\ref{sec:dataset}--\ref{sec:inference}, respectively.

\subsection{Collecting concept activation data}
\label{sec:dataset}

Let $E_I: \mathcal{X} \rightarrow \mathcal{I}$ denote the image encoder and $E_T: \mathcal{Y} \rightarrow \mathcal{T}$ denote the text encoder of CLIP, which map input images and texts into a unified latent representation space. Here, $\mathcal{X} = \mathbb{R}^{H \times W \times 3}$ represents the image space, and $\mathcal{Y}$ denotes the text space. The shared latent space is $\mathcal{I} = \mathcal{T} = \mathbb{R}^{d}$, where $d$ is the dimensionality.  
We denote the training dataset by $D = \{(x_i, y_i)\}$, where $y_i$ is the text associated with image $x_i$, such as a class name or an image caption. For classification datasets, let $Y=(Y_0,\ldots,Y_{|Y|-1})$ be an ordered list of class labels, and $l_i$ denote the class index of $x_i$.

\noindent \textbf{Concept set generation.}  
Following recent works~\citep{oikarinen2023label, yang2023language, yan2023learning}, we adopt a fully automated pipeline that queries an LLM for each class label $y \in Y$ to generate a candidate concept set $C_y$, with final concept set $C = \bigcup_{y \in Y} C_y$, reducing annotation cost and avoiding reliance on scarce datasets with human-defined concepts. For zero-shot classification, we leverage the task-agnostic concept set from SpLiCE~\citep{bhalla2024interpreting}.

\begin{figure}[t]
  \centering
  \includegraphics[width=\linewidth]{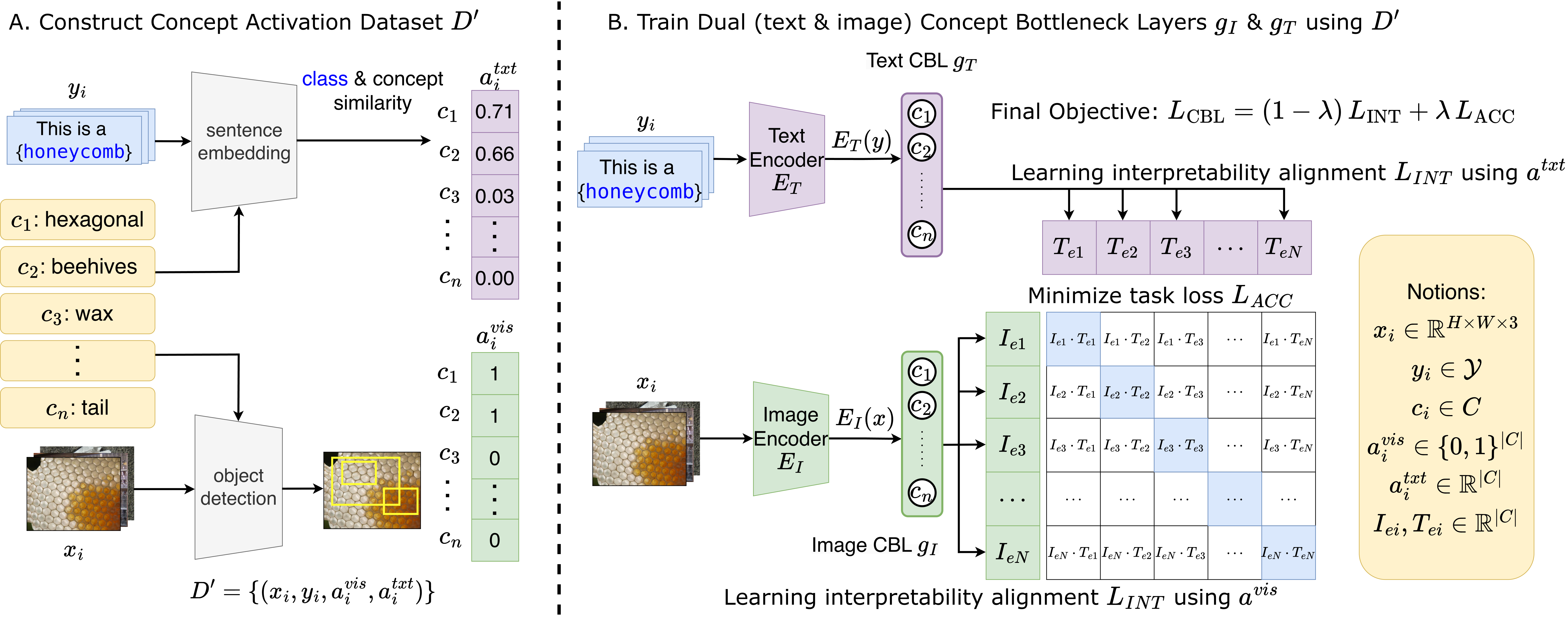}
  \caption{
  Overview of MM-CBM.
  (Left) \textbf{A.} Extracting high-quality concept annotations for each modality. 
  (Right) \textbf{B.} Using an auxiliary dataset to train dual CBLs, jointly optimizing interpretability alignment and task performance. For notational convenience, $n=|C|$ denotes the total number of concepts in this figure, and $N$ denotes the number of image--text pairs in a training batch.
  }
  \label{fig: Overview of model}
\end{figure}

\noindent \textbf{Collecting concept labels.}  
With the candidate concept set $C$ in place, we construct a concept activation dataset $D' = \{(x_i, y_i, a^{\mathrm{vis}}_i, a^{\mathrm{txt}}_i)\}$ by augmenting each image-text pair with two additional labels as shown in Fig.~\ref{fig: Overview of model}A (left):  
\begin{itemize}
    \item $a^{\mathrm{vis}}_i \in \{0, 1\}^{|C|}$, a binary vector indicating which concepts appear in the image $x_i$,
    \item $a^{\mathrm{txt}}_i \in \mathbb{R}^{|C|}$, quantifies how strongly each concept in $C$ relates to the text label $y_i$.  
\end{itemize}

\noindent To equip the model with interpretable supervision, we first extract binary concept labels for each image based on OWLv2's open-source object detection results~\citep{minderer2024scaling}. The image concept label $[a^{\mathrm{vis}}_{i}]_j$ for concept $c_j$ is defined as:
\begin{equation}
\label{Eq:ObjectDetectionLabel}
    [a^{\mathrm{vis}}_{i}]_j = \begin{cases}
        1, & \text{if concept $c_j$ is detected in image $x_i$},\\
        0, & \text{otherwise}.
    \end{cases}
\end{equation}

For each training image $x_i$, we prompt OWLv2 with the class-specific concept set $C_{Y_{l_i}}$. The model predicts a set of bounding boxes $B = \{(b, f, c)\}$, where $b$ represents the box coordinates, $f$ is the confidence score, and $c \in C_{Y_{l_i}}$ is the detected concept. If the confidence $f$ exceeds a predefined threshold $T$, we consider the concept $c$ to be present in image $x_i$.

\noindent To compute $a^{\mathrm{txt}}_i$, the text-concept similarity vector, we define each entry $[a^{\mathrm{txt}}_{i}]_j$ as the semantic similarity between the text label $y_i$ and concept $c_j$:
\begin{equation}
    [a^{\mathrm{txt}}_{i}]_j = \operatorname{sim}(y_i, c_j).
\end{equation}
For the similarity function, we follow the Automatic Concept Scoring (ACS) method from CB-LLM~\citep{sun2024concept}, where similarity is defined as:
\begin{equation}
\label{Eq:TextSimLabel}
    \operatorname{sim}(y_i, c_j) = \mathcal{E}(y_i) \cdot \mathcal{E}(c_j),
\end{equation}
with $\mathcal{E}(\cdot)$ denoting the text embedding generated by a language model. In our implementation, we use the \texttt{all-mpnet-base-v2} model~\citep{wolf2020transformers} as the text encoder.

\subsection{Training dual concept bottleneck layers}
\label{sec:training}
Given the concept activation dataset $D'$, we train a pair of Concept Bottleneck Layers (CBLs): one for identifying concept presence in images, and the other for capturing the association between concepts and textual labels. Our training objective consists of two components: an \textbf{interpretability loss} $L_{\mathrm{INT}}$ and a \textbf{task loss} $L_{\mathrm{ACC}}$, as shown in Fig.~\ref{fig: Overview of model}B (right).

\textbf{Interpretability loss $L_{\mathrm{INT}}$.}
To explicitly align the outputs of the concept bottleneck layers (CBLs) with human-interpretable concepts, we define an interpretability loss that supervises both the image and text sides using binary and soft labels, respectively. Let $g_I: \mathcal{I} \rightarrow \mathbb{R}^{|C|}$ and $g_T: \mathcal{T} \rightarrow \mathbb{R}^{|C|}$ denote the image and text CBLs, respectively, where $|C|$ is the number of concepts. These CBLs project image and text features into a shared concept space. To enforce consistency between predicted concept activations and automatically generated concept annotations in $D'$, we define the interpretability loss as:
\begin{equation}
\begin{aligned}
    L_{\mathrm{INT}} = \frac{1}{|D'|}  \sum_{i=1}^{|D'|} \left[ L_I\big(\hat{a}_i^{\mathrm{vis}}, a_i^{\mathrm{vis}}\big) +      L_T\big(\hat{a}_i^{\mathrm{txt}}, a_i^{\mathrm{txt}}\big) \right], 
\end{aligned}
\end{equation}
where 
$
\hat{a}_i^{\mathrm{vis}} = g_I \circ E_I(x_i), \quad 
\hat{a}_i^{\mathrm{txt}} = g_T \circ E_T(y_i)
$
are the predicted concept activations for the image $x_i$ and text $y_i$, respectively. Here, $E_I$ and $E_T$ are the image and text encoders introduced in Section~\ref{sec:dataset}, and $a^{\mathrm{vis}}_i, a^{\mathrm{txt}}_i \in \mathbb{R}^{|C|}$ are the concept label vectors for image and text, respectively. We adopt binary cross-entropy (BCE) for the image-side loss: 
\begin{equation}
L_I\big(\hat{a}_i^{\mathrm{vis}}, a_i^{\mathrm{vis}}\big)
= -\sum_{k=1}^{|C|} \left[ a_{i,k}^{\mathrm{vis}} \log \sigma(\hat{a}_{i,k}^{\mathrm{vis}}) + (1-a_{i,k}^{\mathrm{vis}}) \log \bigl(1-\sigma(\hat{a}_{i,k}^{\mathrm{vis}})\bigr) \right],
\end{equation}
where \(\sigma(\cdot)\) is the sigmoid function, and we use negative cosine-cubed similarity~\cite{oikarinen2023label} for the text-side loss:
\begin{equation}
\begin{aligned}
L_T\big(\hat{a}_i^{\mathrm{txt}}, a_i^{\mathrm{txt}}\big)
= -
\frac{
    (\hat{a}_i^{\mathrm{txt}})^3 \cdot (a_i^{\mathrm{txt}})^3
}{
    \left\| (\hat{a}_i^{\mathrm{txt}})^3 \right\|_2 \, \left\| (a_i^{\mathrm{txt}})^3 \right\|_2
}
,
\end{aligned}
\end{equation}
reflecting the categorical and continuous nature of the constructed concept labels.

\textbf{Task loss $L_{\mathrm{ACC}}$.}
To preserve discriminative performance on downstream tasks, we introduce a task-specific classification loss based on the representations in the concept space. For notational convenience, we write the predicted image and text concept representations as $I_e=\hat{a}^{\mathrm{vis}}$ and $T_e=\hat{a}^{\mathrm{txt}}$, omitting the sample index.

To ensure interpretability and promote sparsity, i.e., encouraging the prediction to rely on a small subset of semantically meaningful concepts, we draw inspiration from the \textit{number of effective concepts} (NEC)~\citep{srivastavavlg}. Specifically, the similarity between $I_e$ and $T_e$ is computed as the sum of the top-$n$ responding dimensions over element-wise products. To further enhance interpretability and suppress the influence of negatively activated concepts, we set all negative elements in the concept vectors to zero before computing the similarity: $I_e^+ = \operatorname{ReLU}(I_e)$ and $T_e^+ = \operatorname{ReLU}(T_e)$. More details are provided in Appendix~\ref{Ap: enhance}. The semantic consistency between the image and text is then defined as:
\begin{equation}
\label{Eq: logits}    
z_j = \left(\frac{\sum \text{top-}n (I_e^+ \odot T_{e,j}^+)}{\|I_e^+\|_2\|T_{e,j}^+\|_2}\right) \times \exp(\tau),
\end{equation}
where $z_j$ denotes the similarity score (logits) of the image $x$ with respect to the class or query $y_j$, $T_{e,j}=g_T(E_T(y_j))$, and $\odot$ denotes element-wise multiplication. Here, $j$ indexes candidate classes or text queries, with $y_j=Y_j$ for classification, and $\tau$ is a temperature parameter. In this case, the classification loss is expressed as:
\begin{equation}
    L_{\mathrm{ACC}} = L_{\mathrm{CE}}(\mathbf z,\ l).
\end{equation}
Here, $\mathbf z=(z_0,\ldots,z_{|Y|-1})$ contains the logits for all candidate classes, $L_{\mathrm{CE}}$ denotes cross-entropy loss, and $l=l_i$ is the ground-truth class index of image $x_i$.

\textbf{Final objective.} 
To jointly optimize interpretability and task performance, we integrate the interpretability loss and discriminative loss into a unified objective. Notably, our model can also be trained without ground-truth labels, achieving classification accuracy comparable to CLIP; see Appendix~\ref{Ap: KD} for details. This combined loss function enables the model to learn concept-aligned representations while maintaining strong classification performance:
\begin{equation}
\label{Eq: final_loss}
    L_{\mathrm{CBL}} = (1 - \lambda)\, L_{\mathrm{INT}} + \lambda\, L_{\mathrm{ACC}},
\end{equation}
where $\lambda \in [0,1]$ controls the trade-off between interpretability and task accuracy.

\subsection{Performing inference at test time}
\label{sec:inference}

During inference, given any image $x^{\mathrm{test}}$ and text query $y$, the model outputs two modality-specific concept embeddings: $I_e=g_I \circ E_I(x^{\mathrm{test}})$ and $T_e=g_T \circ E_T(y)$, respectively. Their alignment score is computed using Eq.~\ref{Eq: logits}. For classification, we compare the image with all candidate class queries $y_j=Y_j$ and predict the class index $\hat{l}=\arg\max_{j\in\{0,\ldots,|Y|-1\}}z_j$. For image retrieval, we fix the text query and rank candidate images by their alignment scores.

Furthermore, because both embeddings are represented in human-interpretable concept dimensions, each prediction can be explicitly explained by inspecting the concept-wise contributions: 

\begin{equation}
\mathbf{w}_j=\frac{I_e^+\odot T_{e,j}^+}{\|I_e^+\|_2\|T_{e,j}^+\|_2}\exp(\tau),
\qquad z_j=\sum \text{top-}n(\mathbf{w}_j).
\end{equation}

Here, each entry of $\mathbf{w}_j$ corresponds to a concept, and the top-$n$ entries contribute to the similarity score. In contrast to traditional CBMs that rely on a linear classifier to associate concepts with categorical labels, our text CBL directly generates concept activations from natural language descriptions. This not only simplifies the inference pipeline but also enables flexible support for open-vocabulary text inputs, making the post-bottleneck scoring process of MM-CBM fully inspectable, as illustrated in Fig.~\ref{fig: Overview of model}B.

\section{Experiment}
\label{sec:experiment}

Section~\ref{setup} describes the experimental setup. Section~\ref{result} compares MM-CBM to existing CBMs and the CLIP backbone under both fine-tuned and zero-shot settings. Section~\ref{ablation} examines the impact of the interpretability enhancement techniques introduced in Section~\ref{sec:training} and Appendix~\ref{Ap: enhance}. Section~\ref{intervention} investigates how manual interventions can be used to correct certain misclassifications and trace the underlying causes. Section~\ref{interp} presents quantitative interpretability evaluations through human and VLM-based assessments.

\subsection{Experimental setup}
\label{setup}

\textbf{Datasets}: We conduct experiments on seven datasets spanning different types of visual recognition tasks: (1) \textbf{General image classification}: CIFAR-10, CIFAR-100~\citep{krizhevsky2009learning}, and ImageNet~\citep{deng2009imagenet}; (2) \textbf{Fine-grained classification}: Food-101 (Food)~\citep{bossard2014food}, CUB~\citep{wah2011caltech}, and Oxford-IIIT Pets (OxfordPets)~\citep{parkhi2012cats}; (3) \textbf{Texture classification}: Describable Textures Dataset (DTD)~\citep{cimpoi2014describing}. Additionally, we trained MM-CBM on the multimodal dataset CC12M~\citep{changpinyo2021conceptual} as the source dataset for cross-dataset zero-shot transfer. We follow the standard train/test splits, as detailed in Appendix~\ref{Ap: exp config}, and use classification accuracy as the evaluation metric.

\noindent \textbf{Baselines}: We compare MM-CBM with four interpretable baselines: LF-CBM~\citep{oikarinen2023label}, LaBo~\citep{yang2023language}, LM4CV~\citep{yan2023learning}, and VLG-CBM~\citep{srivastavavlg}. We also compare against the CLIP backbone under both zero-shot and target-dataset adaptation settings.

\noindent \textbf{Implementation.}
We set the trade-off parameter to $\lambda=0.2$ throughout. For comparison with prior CBMs, we use CLIP-RN50 as the common backbone, while all other experiments use CLIP-ViT-L/14. The pretrained image and text encoders are frozen in all MM-CBM experiments, and only the dual CBLs and the temperature parameter are learned. We use Adam~\citep{kingma2014adam} for optimization. The main experiments use zero targets for unqueried concepts in the visual BCE loss; an alternative masked BCE formulation and its comparison with current approach are discussed in Appendix~\ref{app:unannotated_concepts}.

In the \textbf{target-dataset adaptation setting (Finetuned)}, MM-CBM uses dataset-specific concept vocabularies generated by \texttt{gpt-3.5-turbo-instruct}, while the CLIP baseline is evaluated using linear probing. In the \textbf{zero-shot transfer setting}, MM-CBM is trained once on CC12M using a fixed task-agnostic SpLiCE vocabulary~\citep{bhalla2024interpreting} of 15,000 concepts and is then evaluated directly on unseen target datasets. No target-dataset training images, ground-truth labels, concept annotations, target-specific vocabulary construction, or parameter fitting are used. Target class names are used only at inference to construct textual candidate queries. In this setting, we use all non-negative concept dimensions rather than NEC$=5$. Detailed protocols and dataset-specific configurations are provided in Appendix~\ref{Ap: exp config}.

\begin{table}[h]

\centering
\caption{Comparison with other CBMs on ANEC-5 using CLIP RN50. Best results for each benchmark are in \textbf{bold}; second-best are \underline{underlined}.}

\begin{tabular}{lccccc}
\toprule
{Method} & \multicolumn{5}{c} {Dataset}\\
    \cmidrule(lr){2-6}
{ANEC=5} & {CIFAR10} & {CIFAR100} & {CUB} & {ImageNet} & {Average}      \\  \midrule 
{LF-CBM}\citep{oikarinen2023label} & 84.05 & 56.52 & 31.35  & 52.88 &  56.20   \\ 
{LM4CV}\citep{yan2023learning}   & 53.72 & 14.64 & 3.63& 3.77 & 18.94 \\
{LaBo}\citep{yang2023language} & 78.69 & 44.82 & 41.97 & 24.27 & 47.44 \\ 
{VLG-CBM}\citep{srivastavavlg} & \underline{88.55} & \underline{65.73} & \textbf{60.38} & \underline{59.74} &  \underline{68.60} \\ 
\textbf{MM-CBM(Ours)} & \textbf{88.69} & \textbf{66.34} & \underline{58.79} & \textbf{70.49} & \textbf{71.08} \\
\bottomrule
\end{tabular}

\label{table: acc_cbm}

\end{table}
\begin{table}[h]

\centering

\caption {Test accuracy comparison with black-box CLIP on zero-shot and finetuned settings.}

\begin{tabular}{lccccccc}
\toprule
    {Method} & \multicolumn{7}{c} {Dataset}\\
    \cmidrule(lr){2-8}
    {} & CIFAR-10 & CIFAR-100 & CUB & Food & OxfordPets & DTD & ImageNet \\
    \midrule
    \textit{Zero-shot}\\
    {CLIP ViT-L/14} & 96.2 & 77.9 & 62.3 & 92.9 & 93.5 & 55.3 & 75.3\\
    {\textbf{MM-CBM(Ours)}} & 95.3 & 75.7 & 36.3 & 84.3 & 76.6 & 50.2 & 66.1 \\
    \midrule
    \textit{Finetuned}\\
    {CLIP linear probe} & 98.0 & 87.5 & 84.5 & 95.2 & 95.1 & 82.1 & 83.9 \\
    {\textbf{MM-CBM(Ours)}} & 97.5 & 84.1 & 75.0 & 93.4 & 91.2 & 75.6 & 81.9 \\
\bottomrule
\end{tabular}

\label{table: acc_clip}
\end{table}

\subsection{Results}
\label{result}

\paragraph{Comparison with existing CBMs:} Table~\ref{table: acc_cbm} shows accuracy under $\text{NEC} = 5$ (ANEC-5). MM-CBM achieves performance comparable to the strongest baseline, VLG-CBM, and surpasses others by over 10\% accuracy on ImageNet. This suggests that MM-CBM benefits most from the rich semantic knowledge embedded in the CLIP backbone, particularly on large-scale datasets. Additional results compared with other CBMs under the zero-shot setting are provided in the Appendix~\ref{Ap: exp config}.

\paragraph{Comparison with CLIP backbone:} Table~\ref{table: acc_clip} compares MM-CBM with CLIP under both fine-tuned and zero-shot settings. 
Across seven datasets, MM-CBM achieves performance comparable to both CLIP’s linear-probe and zero-shot settings. 
A substantial gap remains on CUB in the zero-shot setting: MM-CBM achieves 36.3\% accuracy compared with 62.3\% for CLIP. This gap may partly reflect limited coverage of discriminative bird attributes in the fixed, task-agnostic concept vocabulary, highlighting a limitation of fine-grained zero-shot transfer. Nevertheless, the performance remains non-trivial, indicating that MM-CBM can still generalize reasonably well even under challenging and attribute-rich conditions.

Furthermore, the image retrieval results in Table~\ref{table:retrieval} demonstrate that MM-CBM preserves task effectiveness while offering explicit, concept-level interpretability. The performance gap relative to black-box CLIP remains under 5\%, suggesting that MM-CBM retains most of CLIP’s retrieval capability while providing transparent and human-understandable reasoning.

\begin{table}[t]
\small
\begin{center}
\caption{Zero-shot retrieval comparison with black-box CLIP.}
\begin{tabular}{lcccccccccccc}
\toprule
& \multicolumn{6}{c}{Text Retrieval} & \multicolumn{6}{c}{Image Retrieval} \\
& \multicolumn{3}{c}{Flickr30k} & \multicolumn{3}{c}{MSCOCO} & \multicolumn{3}{c}{Flickr30k} & \multicolumn{3}{c}{MSCOCO} \\
& R@1 & R@5 & R@10 & R@1 & R@5 & R@10 & R@1 & R@5 & R@10 & R@1 & R@5 & R@10 \\
\midrule
CLIP & 85.3 & 97.4 & 99.2 & 56.3 & 79.3 & 86.6 & 64.9 & 87.2 & 92.0 & 36.6 & 61.1 & 71.1 \\
MM-CBM &  71.7 & 92.5 & 96.8 & 41.9 & 67.8 & 78.1 & 61.8 & 85.1 & 90.8 & 32.9 & 58.5 & 68.9 \\
\bottomrule
\end{tabular}
\label{table:retrieval}
\end{center}
\end{table}

\subsection{Ablation study}
\label{ablation}

We assess the effect of the non-negative concept space introduced in Section~\ref{sec:training} and Appendix~\ref{Ap: enhance}. We compare our method against alternatives including sigmoid activation, squaring activations, and a baseline without non-negative processing. Concept sparsity is quantified using the $\ell_1$ norm.

For a non-negative vector $v \in \mathbb{R}^{|C|}$ with $\|v\|_2 = 1$, we have $1 \leq \|v\|_1 \leq \sqrt{|C|}$. Lower $\|v\|_1$ implies higher sparsity, which improves interpretability.

We report the average $\ell_1$ norm of visual ($I_e$) and textual ($T_e$) activations, and average alignment score across validation samples. Table~\ref{table: ablation_l1} shows that our approach yields at least $4\times$ smaller $\ell_1$ norm for visual activations and $2\times$ smaller for text, compared to other methods. Our alignment score also improves by $3\times$, suggesting higher prediction confidence. Importantly, increased sparsity does not degrade accuracy but enhances reliability. Additionally, since visual supervision uses binary targets and text uses real-valued similarity scores, visual concept activations are expected to be sparser. Our method preserves this property, while others reverse it, potentially introducing redundant activations.

\begin{table}[h!]

\caption{Ablation study of non-negative setting. Visual and language correspond to the average $\ell_1$ norm of image and text concept activation; Score means the average highest alignment score, the image and prediction alignment score.}

\centering
\scalebox{1}{
\begin{tabular}{lcccc}
\toprule
{Function} & {Visual} & {Language} & {Alignment} \\
{} & {activation ${\ell_1}
$} $\downarrow$  & {activation ${\ell_1}$
} $\downarrow$ &{score} $\uparrow$ \\
\midrule 
{Sigmoid} & 11.690 & 11.699 & 0.079 \\
{$x^2$} & 58.848 & 23.524 & 0.044 \\ 
{None} & 63.030 & 60.665 & 0.002 \\ 
{\textbf{ReLU}} & \textbf{2.703} & \textbf{5.668} & \textbf{0.271}\\ 
\bottomrule
\end{tabular}
}

\label{table: ablation_l1}
\vspace{0cm}

\end{table}

\subsection {Human Intervention}
\label{intervention}

We evaluate whether the concept bottleneck supports systematic and semantically meaningful interventions. We first conduct a quantitative intervention study on the full ImageNet and CUB test sets. For each image, we use only its positive visual concept-label mask to determine which concept coordinates to edit.

Our primary intervention adds a positive offset to each concept coordinate labeled as present:
\begin{equation}
    I'_{e,c}
    =
    I_{e,c}
    +
    s\delta \cdot \mathbb{I}[a^{\mathrm{vis}}_c = 1],
\end{equation}
where $\delta$ is the 90th percentile of positive visual CBL activations and $s$ controls the intervention strength. After intervention, we re-normalize the visual concept representation and rerun the complete NEC-$5$ classification procedure. A mild intervention with $s=0.5$ achieves the best trade-off between correcting erroneous predictions and preserving previously correct ones. On ImageNet, top-1 accuracy improves from 81.74\% to 87.06\%, correcting 41.08\% of the originally incorrect predictions while changing only 2.67\% of the originally correct predictions. This results in a net gain of 2,660 correctly classified samples. On CUB, top-1 accuracy improves from 76.46\% to 83.16\%, correcting 60.67\% of the original errors while changing 9.92\% of previously correct predictions, resulting in a net gain of 388 samples.

To verify that the improvement depends on concept semantics rather than simply increasing a fixed number of coordinates, we construct a random permutation control that preserves the number of positive concepts per image and the global concept-frequency distribution while disrupting the correspondence between concept labels and CBL coordinates. Under this control, top-1 accuracy is 81.08\% on ImageNet and 53.54\% on CUB, substantially below the semantically aligned intervention. These results indicate that the intervention benefit relies on the learned alignment between named concepts and CBL coordinates.

We additionally find that stronger edits introduce substantially greater collateral damage. Full results for alternative intervention operators and intervention strengths are provided in Appendix~\ref{app:intervention}.

\begin{figure}[h]
  \centering
  \includegraphics[width=\linewidth]{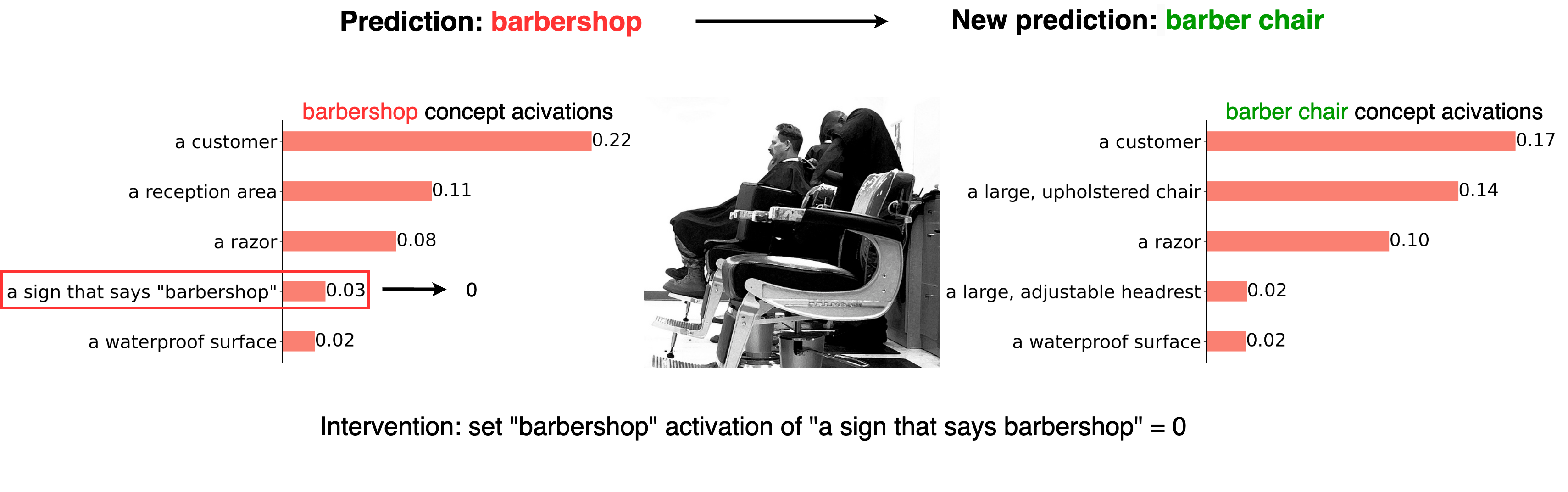}
  \caption{A sample of correcting model prediction by deleting the wrong concept.}
  \label{fig:intervention}
\end{figure}

\paragraph{Qualitative expert-guided interventions.}
Beyond the systematic image-side intervention above, MM-CBM also supports targeted edits based on inspection of the text-side concept responses. These edits illustrate how specific prediction errors can be traced to individual concept coordinates and corrected using expert knowledge. Fig.~\ref{fig:intervention} shows a misclassification where the model predicts ``barbershop'' due to a strong activation of the concept ``a sign that says barbershop,'' which is not visually present. This can be corrected by setting $T_{e[\text{pred, concept}]} = 0$.

\subsection{Interpretability result}
\label{interp}

\textbf{Human study.} To evaluate the interpretability of our model, we conducted a user study following a similar setting to Label-free CBM~\citep{oikarinen2023label}. We compared two variants of MM-CBM (zero-shot and fine-tuned) against SpLiCE~\citep{bhalla2024interpreting}. For each method, participants were shown 100 randomly sampled ImageNet images with the correct label. For every sample, we displayed the top five contributing concepts and their corresponding weights, and asked users to rate ``which explanation is better'' on a Likert scale from 1 to 5.

Across both evaluation settings, users consistently preferred the explanations produced by MM-CBM. Participants further provided qualitative feedback: in the zero-shot setting, MM-CBM offered descriptions that were more relevant to the image in 55.33\% of cases and provided more informative explanations in 56.25\% of cases. In the fine-tuned setting, these preferences increased to 64.52\% and 66.57\%, respectively, indicating that participants more frequently preferred MM-CBM explanations in terms of perceived relevance and informativeness. Our study was deemed Exempt from IRB approval by the IRB review board at our institution.

\begin{figure}[ht]
  \centering
  \includegraphics[width=0.45\linewidth]{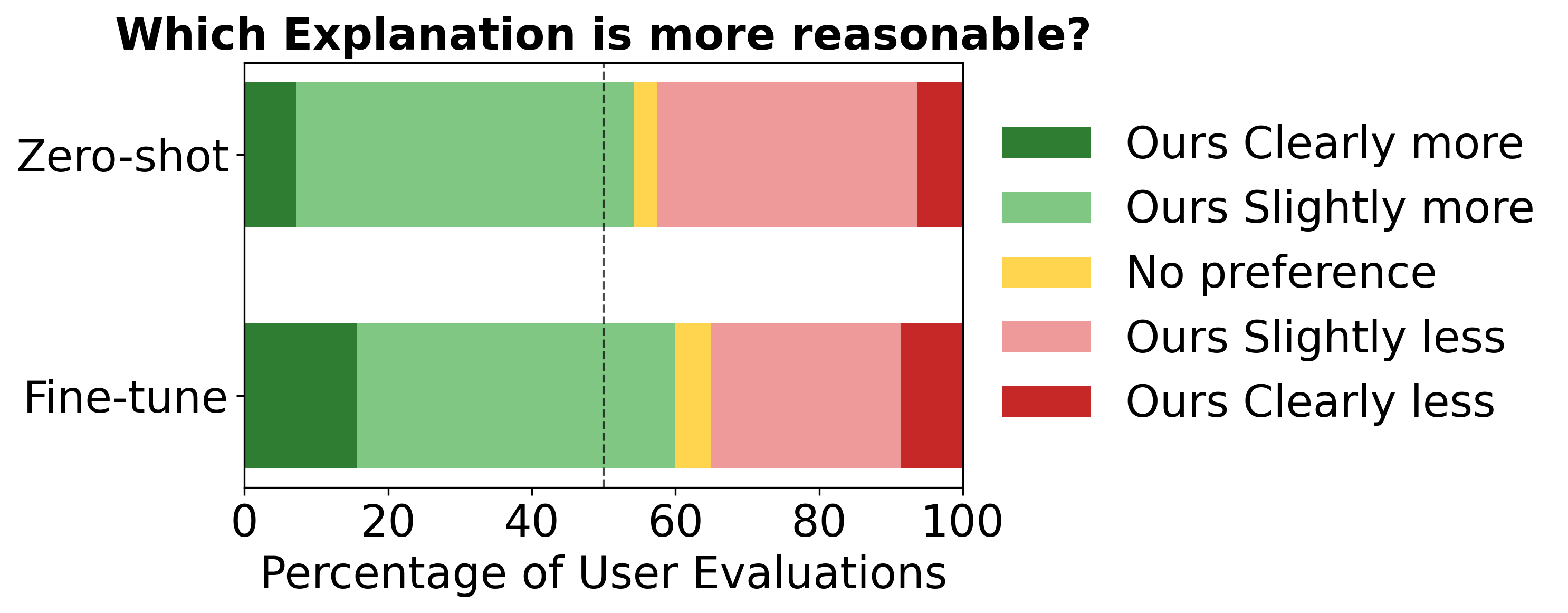}
  \caption{
    Interpretability result
  }
  \label{fig: Interp result}
\end{figure}

\textbf{Comparison with VLMs} Vision-Language Models (VLMs) are highly capable of understanding the overall semantics of images and generating natural language explanations. This appears similar to the goal of CBMs, so we directly compared explanations from our MM-CBM (ImageNet) with those from VLMs.

Specifically, we prompted each model with the template: ``Why is this image categorized as \{cls\}'', and collected 5,000 explanation pairs from imagenet dataset. To evaluate which explanation contained more informative visual concepts, we leveraged the VQA capabilities of VLMs themselves by asking: ``Which description has more informative visual concepts in this image?'' Notably, to reduce model-specific bias and avoid self-preference in scoring, we separated the roles of evaluator and competitor across models, using different VLMs to act as the ``judge'' and the ``explainer.'' In our experiments, we used \texttt{LLaVA-v1.5-7B}~\citep{liu2023visual} and \texttt{Llama-3.2-11B-Vision-Instruct}~\citep{touvron2023llama}, alternating their roles to ensure fairness and robustness. MM-CBM explanations were preferred over LLaVA 1.5 in \textbf{4,433} (88.7\%) out of 5,000 cases, and over Llama 3.2 in \textbf{3,292} (65.8\%) cases. These results indicate that, under this evaluation protocol, MM-CBM explanations are more frequently judged to contain informative visual concepts than the explanations generated by the evaluated VLMs.

\section{Case study: image retrieval}
\label{sec:usecase}

\begin{figure*}[h!]
  \centering
  \includegraphics[width=\linewidth]{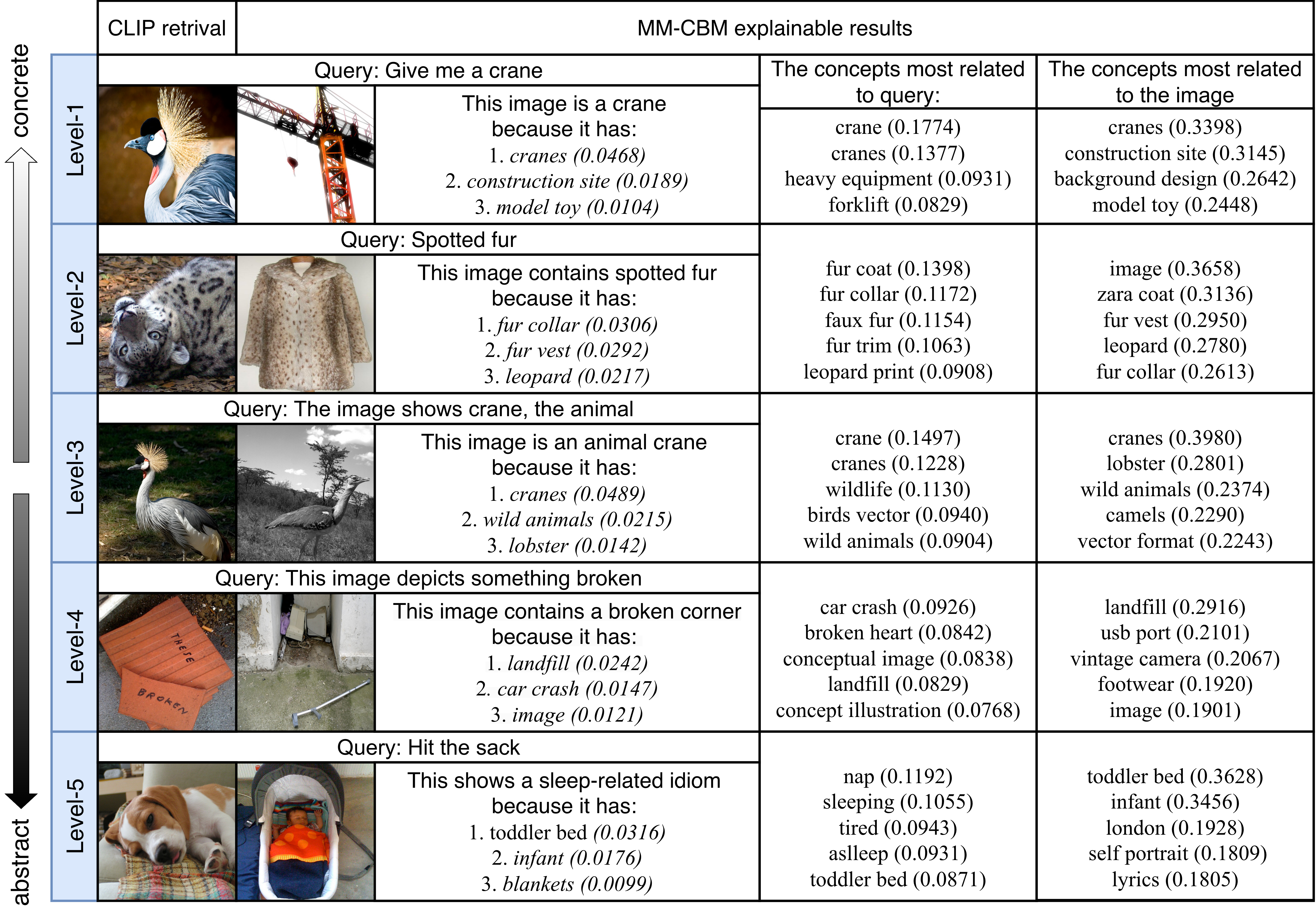}
  \caption{Image Retrieval on five different levels of queries. By incorporating the textual CBL, the model can infer contextual meanings: in Level-3, the query ``crane the animal'' successfully retrieves the bird instead of the mechanical object; in Level-4, the term ``broken'' evokes semantically related scenes; and in Level-5, the model captures contextual distinctions between the literal and figurative meanings of polysemous phrases. For readability, we show the top three concepts ranked by their contributions to the image--text alignment score, together with the top five concepts ranked by their normalized activations in each modality.}
  \label{fig:imageretrieval}
\end{figure*}

We replace the fixed linear classifier with a text encoder, enabling flexible and unrestricted text inputs. In this section, we evaluate our model through an image retrieval task: given arbitrary text, the model selects the image with the highest alignment score. This evaluation assesses the model’s semantic consistency and generalization ability. To systematically analyze interpretable reasoning in image retrieval, we define five levels of textual queries:
\begin{itemize}
    \item \textbf{Level-1: Ground-truth label queries} – Direct retrieval using exact class labels (e.g., \textit{crane}, \textit{popsicle}, \textit{uniform}).
    \item \textbf{Level-2: Concept-based queries} – Retrieval based on key concepts (e.g., \textit{striped fur}, \textit{spotted fur}, \textit{uniformed fur}), allowing us to test fine-grained concept understanding.
    \item \textbf{Level-3: Hybrid label–concept queries} – Queries that combine class labels and specific concepts (e.g., \textit{crane the animal}, which resolves the ambiguity of the polysemous word ``crane,'' or \textit{cat with striped fur}, which resembles a \textit{tiger}).
    \item \textbf{Level-4: Out-of-distribution queries} – Texts containing unseen labels or novel concepts not present in the training set (e.g., \textit{broken}, which typically describes a state rather than a class in standard classification tasks).
    \item \textbf{Level-5: Polysemous or abstract queries} – Phrases involving semantic ambiguity, idiomatic expressions, or abstract meanings. (e.g., \textit{hit the sack} meaning ``go to sleep'', \textit{Break a leg}, which literally means ``injure a leg'' but figuratively conveys ``good luck'' in theatrical slang
\end{itemize}

This hierarchical evaluation setup helps validate the model’s semantic alignment and its ability to generalize beyond predefined labels or fixed concept sets. We conduct all experiments using the model trained on CC12M, and representative examples for each query level are shown in Fig.~\ref{fig:imageretrieval}. The retrieval results demonstrate that our model’s semantic understanding is largely consistent with both human judgment and the backbone model, while additionally providing an interpretable account of how such semantic decisions are formed.

We further present additional cases in the Appendix~\ref{Ap: ir_sup} Fig.~\ref{fig:imageretrieval_failurecase}, particularly for Level-4 and Level-5 queries, where the showcases are less accurate or even incorrect compared to the original black-box model. In these cases, the text modality often captures concepts that are semantically similar to the query. However, since our visual concept labels are constructed via object detection, the visual branch is inherently less sensitive to abstract semantics and instead responds more strongly to concrete objects (Appendix~\ref{app:relational_annotation}). Moreover, as shown in Table~\ref{table: ablation_l1}, visual activations are generally higher than textual ones, leading the final explanations to be dominated by visual signals. This imbalance helps explain the observed failure cases. In contrast, for more concrete queries (Level-1 to Level-3), concept activations are consistently stronger, indicating that the model is more confident when handling explicit and visually grounded semantics.
\section{Conclusion}
\label{sec:conclusion}

In summary, we present MM-CBM, a simple yet effective framework that turns vision–language foundation models into interpretable, concept-based predictors. MM-CBM provides fully inspectable post-bottleneck scoring and supports concept-mediated classification, zero-shot transfer, image--text retrieval, and intervention without relying on an additional non-concept classification path. Across the evaluated settings, MM-CBM outperforms prior CBM baselines under ANEC-5 and remains competitive with black-box CLIP. Our human and VLM preference studies show that MM-CBM explanations are frequently judged to be more relevant and informative, while the sufficiency, comprehensiveness, and systematic intervention evaluations provide evidence that the highlighted concepts are decision-relevant and can be used to inspect and modify model predictions. These findings position MM-CBM as a practical concept-mediated alternative for inspectable multimodal prediction and suggest its potential as a trustworthy building block for multimodal applications.
\section*{Broader Impacts}
\label{sec:conclusion}

MM-CBM improves the interpretability of vision–language foundation models by grounding predictions in human-understandable concepts. However, it may be subject to misuse, whereby harmful or discriminatory concepts are injected during the training process, potentially leading to biased or undesirable outputs. In addition, since the training data is labeled based on black-box foundation models, the learned concepts may not fully align with human understanding, introducing further risks. To mitigate these issues, MM-CBM enables human intervention during inference and supports error tracing, facilitating more transparent diagnosis and partial control over model behavior.
\section*{Acknowledgments}

The authors are partially supported by National Science Foundation under Grant No. 2313105, 2430539, and Intel Rising Star Faculty Award. The authors would like to thank anonymous reviewers for valuable feedback to improve manuscript.

\bibliography{main}
\bibliographystyle{tmlr}

\clearpage
\appendix

\clearpage
\appendix

\startcontents[appendices]
\section*{Appendix Contents}
\printcontents[appendices]{l}{1}{\setcounter{tocdepth}{2}}

\clearpage

\counterwithin{figure}{section}
\counterwithin{table}{section}
\counterwithin{equation}{section}

\section{Method and Experimental Setup}
\subsection{Problem Formulation}
\label{Ap: problem formulation}

Given an input image $I$ and a text description $T$, we aim to estimate the conditional probability $p(T|I)$, i.e., the likelihood that $T$ correctly describes $I$. 
We assume that both modalities share an intermediate concept space ${c_i}_{i=1}^K$, where each concept $c_i$ captures a semantic attribute relevant to both image and text. Strict equality holds only if the concept set $C$ fully covers the semantic space and the concepts are mutually disjoint; in practice, we adopt this as an approximation.

Under this assumption, the image-text relation can be decomposed through the latent concepts as follows:
\begin{equation}
    p(T|I) \approx \sum_i p(T, c_i|I).
\end{equation}
By applying the product rule of probability, we have
\begin{equation}
\begin{cases}
    p(T, c_i|I) = p(T|c_i, I)\,p(c_i|I), \\
    p(T|c_i, I) = p(T|c_i),
\end{cases}
\end{equation}
where the second equality assumes that, given the concept $c_i$, the text $T$ is conditionally independent of the image $I$.  
Substituting into the previous equation yields
\begin{equation}
    p(T|I) \approx \sum_i p(T|c_i)\,p(c_i|I).
\end{equation}
Next, by applying Bayes’ theorem, the term $p(T|c_i)$ can be expressed as
\begin{equation}
    p(T|c_i) = \frac{p(c_i|T)\,p(T)}{p(c_i)}.
\end{equation}
The concept prior $p(c_i)$ can be marginalized over the image distribution:
\begin{equation}
    p(c_i) = \sum_j p(c_i|I_j)\,p(I_j).
\end{equation}
Combining the above relations gives
\begin{equation}
    p(T|I) \approx \sum_i p(c_i|I)\,\frac{p(c_i|T)\,p(T)}{\sum_j p(c_i|I_j)\,p(I_j)}.
\end{equation}
When assuming uniform priors for $p(I)$ and $p(T)$, the proportional relation becomes
\begin{equation}
    p(T|I) \propto \sum_i p(c_i|I)\,\frac{p(c_i|T)}{\sum_j p(c_i|I_j)}.
\end{equation}
This formulation provides a probabilistic bridge between the image and text modalities through their shared concept distributions $p(c_i|I)$ and $p(c_i|T)$, which can be directly predicted by our CBL layers, allowing us to evaluate the semantic consistency between $I$ and $T$ via their inferred concept activations. Our work presents an initial implementation of this modeling approach. Future research could further enhance interpretability and task performance by improving the quality of the concept set, e.g., expanding its semantic coverage and reducing overlap, and by refining the conditional concept distribution model to better approximate the true underlying distribution.

\subsection{Strategies for Enhancing Interpretability}
\label{Ap: enhance}

In this section, we introduce three strategies designed to enhance the interpretability of our multimodal CBM model.

\paragraph{Generation of rich textual information.}  
In many vision-language datasets, there exists a significant imbalance between the number of images and the granularity of their associated textual labels, where hundreds or even thousands of images may share the same class name. Repeatedly using identical textual inputs during training can introduce undesirable biases and restrict model generalization. To address this issue, we leverage the capabilities of the state-of-the-art multimodal large language model \texttt{Llama 3.2-Vision} to generate diverse, semantically rich label descriptions. Specifically, we prompt the model with the following template: \textit{``If I had to describe this image using only one sentence with the words {class}, it would be: ''} For each class label, we randomly select images belonging to that class and generate at least 50 unique textual descriptions. During training, one of these alternative descriptions is randomly sampled for each iteration, thereby improving diversity in the language modality and reducing overfitting to fixed textual patterns.

\begin{table}[h!]
\centering
\caption{Examples of generated sentence.}
\begin{tabular}{c}
\toprule
{Generated sentence}\\  \midrule 
The image shows a hand holding tench. \\
This is a close-up of a goldfish. \\ 
The image depicts a jay with its wings spread.  \\ 
It would be a smooth newt with a smooth skin.  \\ 
The peafowl is pecking at the ground. \\ 
The macaw is a vibrant and colorful bird. \\ 
The image features a Bluetick Coonhound. \\ 
\bottomrule
\end{tabular}
\label{table:generated_case}
\end{table}

\paragraph{Number of effective concepts (NEC).} 
NEC, originally proposed by~\cite{srivastavavlg}, is a metric that helps prevent information leakage by constraining model reliance on a limited set of semantically meaningful features. We adapt this approach to our multimodal CBM when computing the alignment score between an input image $x_i$ and its corresponding label $y_i$ using interpretable encodings. Specifically, we select the top-$n$ dimensions from the element-wise similarity between image and text encodings and use their sum as the final similarity score:

\begin{equation}
\label{Eq:NEC}
    \text{logits} = \left(\frac{\sum \text{top-}n (I_e \odot T_e)}{\|I_e\|_2\|T_e\|_2}\right) \times e^\tau
\end{equation}

Here, $\odot$ denotes element-wise multiplication. Our method dynamically identifies the top-$n$ most relevant concepts for each image-text pair, making the reasoning process more interpretable and supporting better downstream interventions.

\paragraph{Non-negative concept representation space.}  
In our concept representation space, each dimension reflects the similarity between the input (image or text) and a specific concept. To improve interpretability, we enforce a non-negative constraint on these activations by applying a ReLU function to both image and text embeddings: $I_e^+ = \text{ReLU}(I_e)$ and $T_e^+ = \text{ReLU}(T_e)$. This design improves interpretability in the following three aspects:

\begin{enumerate}
    \item \textit{Disambiguating negative responses.} As discussed by~\cite{sun2024concept}, it is often unclear whether a negative activation implies the negation of a concept or its complete absence. By removing negative values, we avoid this ambiguity.
    
    \item \textit{Amplifying relevant concept activations.} Since similarity computations involve normalization, weak activations in high-dimensional spaces can lead to dilution of important signals. By zeroing out irrelevant (negative) dimensions, we strengthen the contribution of meaningful concepts. In the worst-case scenario when normalized activations are uniformly distributed across all \(|C|\) concepts, each coordinate equals \(1/\sqrt{|C|}\), where $|C|$ is the number of candidate concepts; thus, filtering noise is crucial.
    
    \item \textit{Improving inference reliability and efficiency.} Without non-negativity, the product of two negative activations (from image and text encodings) may yield a misleadingly high similarity score, falsely indicating semantic alignment. Enforcing non-negativity eliminates this issue and also simplifies the computation and sorting steps during inference.
\end{enumerate}

\paragraph{Concept contributions and visualization.}
For visualization, we decompose the final prediction score into individual concept contributions.
For image $x_i$ and text query $y_j$, the contribution of concept $c$ is computed as

\[
w_{i,j,c}=
\frac{
I^+_{e,i,c}T^+_{e,j,c}
}{
\|I^+_{e,i}\|_2\|T^+_{e,j}\|_2
}
e^{\tau}.
\]

Here, $I^+_{e,i}$ and $T^+_{e,j}$ denote the non-negative image and text concept
representations after the Concept Bottleneck Layers. For image-side visualization (the output of Image CBL in Figure~\ref{fig: Comparison of model} and ``The concepts most related to the image'' in the retrieval results), the reported concept values are obtained from the normalized image-side concept representation $I^+_{e,i}/\|I^+_{e,i}\|_2$. Similarly, for text-side visualization (the output of Text CBL in Figure~\ref{fig: Comparison of model} and ``The concepts most related to the query'' in the retrieval results), the reported values are determined based on the normalized text-side concept representation
$T^+_{e,j}/\|T^+_{e,j}\|_2$.

For prediction explanations, the displayed concept values correspond to the decision contributions $w_{i,j,c}$ rather than raw concept activations. These values represent the contribution of each concept to the final image-text similarity score. The final prediction score is computed as the summation of the selected concept contributions:

\[
z_{i,j}
=
\sum_{ c\in top-n_{i,j} } w_{i,j,c}.
\]

Therefore, the displayed concept contributions are directly additive and sum to the reported prediction score under the Top-$n$ inference procedure.

\subsection{Experimental Protocols and Configurations}
\label{Ap: exp config}

\begin{table*}[t!]

\centering

\caption {Test accuracy comparison with CBMs on zero-shot settings. *Since LaBo~\citep{yang2023language} does not provide an official OxfordPets concept set, we conducted evaluation using our own data.}

\begin{tabular}{lccccccc}
\toprule
    {Method} & \multicolumn{7}{c} {Dataset}\\
    \cmidrule(lr){2-8}
    {} & CIFAR-10 & CIFAR-100 & CUB & Food & OxfordPets & DTD & ImageNet \\
    \midrule
    {CLIP ViT-L/14} & 96.2 & 77.9 & 62.3 & 92.9 & 93.5 & 55.3 & 75.3\\
    {LaBo}~\citep{yang2023language} & 82.1 & 43.4 & 16.2 & 54.0 & 13.9* & 38.1 & 37.8\\
    {ALBM}~\citep{zhang2025attribute} & 83.1 & 43.1 & 25.0 & 75.4 & 35.9 & 48.5 & 64.6 \\
    {\textbf{MM-CBM(Ours)}} & \textbf{95.3} & \textbf{75.7} & \textbf{36.3} & \textbf{84.3} & \textbf{76.6} & \textbf{50.2} & \textbf{66.1} \\
\bottomrule
\end{tabular}

\label{table: acc_clip_app}
\end{table*}

We distinguish the following training and evaluation protocols:

\begin{itemize}

    \item \textbf{Comparison with prior CBMs (Table~\ref{table: acc_cbm}).}
    We use CLIP-RN50 as the common backbone to ensure consistency with prior
    CBM evaluations. All methods are evaluated under ANEC-5. For MM-CBM,
    the image and text encoders are frozen, while the dual CBLs and the
    temperature parameter are trained on the target dataset. We use
    dataset-specific concept vocabularies generated by
    \texttt{gpt-3.5-turbo-instruct}, with the corresponding vocabulary sizes
    reported in Table~\ref{table:dataset_config}. The baseline concept
    sources and training configurations follow their official implementations
    or reported settings.

    \item \textbf{Target-dataset adaptation (Table~\ref{table: acc_clip}).}
    We use CLIP-ViT-L/14 for both CLIP and MM-CBM. For the CLIP baseline,
    both encoders are frozen and only a linear classification head is trained;
    we therefore refer to this setting as \emph{linear probing}, rather than
    full-model fine-tuning. For MM-CBM, both pretrained encoders remain
    frozen, while the image CBL, text CBL, and temperature parameter are
    optimized using the target-dataset training split. MM-CBM uses
    dataset-specific GPT-generated concept vocabularies,
    $\lambda=0.2$, and NEC$=5$.

    \item \textbf{Cross-dataset zero-shot transfer
    (Table~\ref{table: acc_clip} and Table~\ref{table: acc_clip_app}).}
    We train the dual CBLs on CC12M using CLIP-ViT-L/14 and a fixed
    task-agnostic SpLiCE vocabulary~\citep{bhalla2024interpreting} containing 15,000 concepts. The same
    trained CBLs and concept vocabulary are then kept fixed and evaluated
    directly on all target datasets. No target-dataset training images,
    ground-truth labels, concept annotations, or target-specific parameter
    optimization are used. Target class names are used only to construct
    inference-time text queries through prompt ensembling. In this setting,
    we use all non-negative concept dimensions for similarity computation
    rather than restricting the score to NEC$=5$. The source-domain training
    uses $\lambda=0.2$.

\end{itemize}

The temperature parameter is initialized at
$\tau=0.07$, and all MM-CBM models are optimized using Adam.

\begin{table}[h]
\centering
\caption{
Dataset statistics and dataset-specific MM-CBM configurations for
target-dataset adaptation. Concept counts correspond to GPT-generated
dataset-specific vocabularies. The zero-shot transfer setting instead uses
the fixed 15,000-concept SpLiCE vocabulary~\citep{bhalla2024interpreting} described above.
}
\label{table:dataset_config}
\begin{tabular}{lrrrrrr}
\toprule
Dataset &
Classes &
Train size &
Test size &
Batch size &
Epochs &
Concepts \\
\midrule
CIFAR-10   & 10   & 50,000    & 10,000 & 128 & 50 & 143   \\
CIFAR-100  & 100  & 50,000    & 10,000 & 64  & 50 & 892   \\
CUB        & 200  & 5,994     & 5,794  & 8   & 50 & 370   \\
Food-101   & 101  & 75,750    & 25,250 & 128 & 50 & 768   \\
OxfordPets & 37   & 3,680     & 3,669  & 8   & 50 & 250   \\
DTD        & 47   & 3,760     & 1,880  & 4   & 50 & 689   \\
ImageNet   & 1000 & 1,281,167 & 50,000 & 256 & 12 & 4,751 \\
\bottomrule
\end{tabular}
\end{table}

\section{Concept Vocabulary and Supervision}
\subsection{Concept Vocabulary Generation}
\label{Ap: concept set generation}
Let $C$ denote a fine-grained concept set that semantically explains the images and their corresponding labels in $D$. Such a set can be manually curated by domain experts or automatically generated using large language models (LLMs)~\citep{touvron2023llama, brown2020language}. Following recent studies~\citep{oikarinen2023label, yang2023language, yan2023learning}, we adopt a fully automated approach in which, for each class label $y \in Y$, an LLM is queried to produce a candidate concept set $C_y$. Following the prompt design of label-free CBM~\citep{oikarinen2023label}, we use the following three prompt types:
\begin{itemize}
    \item List the most important features for recognizing something as a \{class\}:
    \item List the things most commonly seen around a \{class\}:
    \item Give superclasses for the word \{class\}:
\end{itemize}
Here, \{class\} refers to the class name in the target classification task. The final concept set is obtained as the union of all class-specific sets:  
\[
C = \bigcup_{y \in Y} C_y.
\]  
We further refine $C$ using the filtering strategy proposed in label-free CBM~\citep{oikarinen2023label}, with the following steps:  
\begin{enumerate}
    \item \textbf{Concept length:} Discard concepts exceeding 30 characters to maintain simplicity and interpretability.
    \item \textbf{Similarity to target classes:} Remove concepts overly similar to target class names, as they undermine the explanatory role of the CBM. Similarity is measured via cosine similarity in a joint text embedding space, combining features from the CLIP ViT-B/16 text encoder and the all-mpnet-base-v2 sentence encoder. Concepts with similarity greater than $0.85$ to any target class are excluded.
    \item \textbf{Redundancy removal:} Eliminate duplicate or near-synonymous concepts to ensure diversity in the bottleneck layer. Using the same embedding space, any concept with cosine similarity above $0.9$ to an already retained concept is removed.
\end{enumerate}
This automated generation and filtering process substantially reduces the reliance on manual annotation while enabling scalable construction of rich concept sets, even for datasets lacking human-defined concept annotations.

\paragraph{Generation configuration and reproducibility.}

We generate dataset-specific concept vocabularies using \texttt{gpt-3.5-turbo-instruct}. The decoding parameters are set to a temperature of $0.7$, top-$p$ of $1.0$, and a maximum output length of $256$ tokens, with both the frequency penalty and presence penalty set to zero. A sampling seed was not specified or recorded for the original concept generations, which we disclose as a limitation of the current reproducibility setup.

For each class, each of the three prompt types is queried twice, resulting in six LLM calls per class in each generation run. The number of concepts returned by the model is not explicitly fixed. Across five independently generated CIFAR-100 vocabularies, the pipeline produces an average of $26.42$ unique concepts per class before filtering and retains an average of $17.49$ concepts per class after filtering.

All prompt templates contain fixed few-shot demonstrations. The important-feature prompt uses ``goldfish'' and ``beerglass'' as demonstrations, while the superclass and common-surroundings prompts use ``tench'' and ``beer glass.'' The complete prompt templates are provided below.

\clearpage

\begin{small}
\begin{verbatim}
Important features:
List the most important features for recognizing something as a "goldfish":

-bright orange color
-a small, round body
-a long, flowing tail
-a small mouth
-orange fins

List the most important features for recognizing something as a "beerglass":

-a tall, cylindrical shape
-clear or translucent color
-opening at the top
-a sturdy base
-a handle

List the most important features for recognizing something as a "{class}":

Superclasses:
Give superclasses for the word "tench":

-fish
-vertebrate
-animal

Give superclasses for the word "beer glass":

-glass
-container
-object

Give superclasses for the word "{class}":

Common surroundings:
List the things most commonly seen around a "tench":

-a pond
-fish
-a net
-a rod
-a reel
-a hook
-bait

List the things most commonly seen around a "beer glass":

-beer
-a bar
-a coaster
-a napkin
-a straw
-a lime
-a person

List the things most commonly seen around a "{class}":
\end{verbatim}
\end{small}

To evaluate generation stability, we repeat the complete concept-generation pipeline five times on CIFAR-100 and compute semantic Jaccard similarity for all
\[
\binom{5}{2}=10
\]
pairs of generated vocabularies. Concepts are embedded using \texttt{all-mpnet-base-v2}. For each pair of runs, concepts are greedily matched one-to-one when their cosine similarity exceeds $0.75$. The matched pairs define the semantic intersection, while unmatched concepts contribute to the union. Across all ten pairwise comparisons, the average semantic Jaccard similarity is $0.48$. This result indicates moderate semantic overlap across stochastic generations, while also revealing non-negligible variation in lexical choice and concept granularity.

{

\subsection{Handling Unannotated Concepts}
\label{app:unannotated_concepts}

\paragraph{Assign-zero and masked supervision.}
In the original implementation, OWLv2 is queried with the class-specific concept subset for each training image, and concept dimensions outside this subset receive zero targets in the visual BCE loss. This assign-zero configuration is used for the main experimental results. However, an unqueried concept is unannotated rather than verified absent, so assigning it a negative target may suppress visually present concepts. We therefore compare assign-zero supervision with masked BCE, which excludes unqueried dimensions from the visual supervision loss while retaining supervision on queried dimensions.

\paragraph{Controlled comparison.}
We compare the two objectives on ImageNet using the same 4,751-concept vocabulary and identical training settings. As shown in Table~\ref{tab:masked_bce}, their Top-1 accuracies differ by 0.47 percentage points. Nevertheless, their learned representations differ: masked BCE produces higher activations on unqueried concepts, with an average of 106 additional unqueried coordinates per image exceeding an activation threshold of 0.5. This increase alone does not establish that the additional activations correctly represent concepts present in the image.

\begin{table}[t]
    \centering
    \caption{ImageNet Top-1 accuracy (\%) under assign-zero and masked BCE supervision. Removing unqueried coordinates is a diagnostic analysis based on the annotation query sets.}
    \label{tab:masked_bce}
    \begin{tabular}{lcc}
        \toprule
        Visual supervision & Original & Unqueried coordinates removed \\
        \midrule
        Assign-zero & 81.91 & 86.46 \\
        Masked BCE & 81.44 & 80.86 \\
        \bottomrule
    \end{tabular}
\end{table}

\paragraph{Diagnostic removal of unqueried concepts.}
Removing unqueried coordinates from the masked model decreases Top-1 accuracy from 81.44\% to 80.86\% and changes 30.63\% of predictions, indicating that these coordinates affect predictions and provide a modest net accuracy benefit in this evaluation. In contrast, removing them from the assign-zero model increases accuracy from 81.91\% to 86.46\%. These results show that similar overall classification accuracy can coexist with substantially different behavior of unqueried coordinates. This annotation-dependent analysis is diagnostic and should not be interpreted as an inference procedure that requires no annotation information.

\paragraph{Annotation reliability and recommendation.}
A human audit covering 151 images and 453 concept--image pairs provides additional evidence that unqueried concepts cannot be assumed absent. Among the audited OWLv2-positive annotations, 90.7\% were judged visually present; among sampled high-activation concepts outside the queried set, 72.9\% were judged present. We therefore recommend masked BCE when concept annotations are incomplete, treating unqueried dimensions as unknown rather than negative. This recommendation avoids imposing negative supervision solely because a concept was not queried.

\subsection{Relational Concept Annotation}
\label{app:relational_annotation}

\paragraph{Scope of detector-based supervision.}
We use OWLv2 to obtain scalable, visually grounded concept supervision. However, object detection emphasizes objects and attributes and may inadequately distinguish activities, interactions, and scene-level context. We investigate this limitation through an annotation-level comparison with GPT-4o.

\paragraph{Evaluation protocol.}
We sample 500 images from the COCO validation set, each accompanied by five human-written captions. Using GPT-4o, we construct caption-derived concepts covering activities/events, functional interactions, social/group relationships, scene context, and global visual states. For each positive phrase, we construct a hard negative that preserves the entities while changing the predicate or state, such as ``dog jumping to catch a frisbee'' versus ``dog resting beside a frisbee.'' An independent caption-only filtering step removes concepts that cannot be verified from static images, yielding 236 positive--negative pairs. During visual evaluation, GPT-4o receives only the image and candidate phrases, without access to the captions or generation process. OWLv2 is evaluated on the same phrase pairs.

\begin{table}[t]
    \centering
    \caption{Annotation-level comparison on 236 relational concept pairs. All values except AUC are percentages.}
    \label{tab:relational_annotation}
    \begin{tabular}{lcc}
        \toprule
        Metric & GPT-4o & OWLv2 \\
        \midrule
        Positive phrase ranked above hard negative & 82.63 & 53.81 \\
        Positive-vs-negative AUC & 0.897 & 0.541 \\
        Positive concept detection rate & 89.41 & 99.58 \\
        Hard-negative false positive rate & 19.49 & 97.03 \\
        \bottomrule
    \end{tabular}
\end{table}

\begin{table}[t]
    \centering
    \caption{Percentage of pairs in which the positive phrase is ranked above its hard negative, for selected concept types.}
    \label{tab:relational_annotation_types}
    \begin{tabular}{lcc}
        \toprule
        Concept type & GPT-4o & OWLv2 \\
        \midrule
        Activity/event & 86.0 & 48.0 \\
        Scene context & 84.6 & 59.0 \\
        Functional interaction & 70.5 & 56.8 \\
        Social/group relation & 85.7 & 57.1 \\
        \bottomrule
    \end{tabular}
\end{table}

\paragraph{Findings and limitations.}
As shown in Table~\ref{tab:relational_annotation}, OWLv2 frequently detects positive phrases but also accepts hard negatives containing the same entities, indicating limited discrimination of the changed relation or state. GPT-4o provides stronger pairwise discrimination, including for activities/events and scene context (Table~\ref{tab:relational_annotation_types}). These findings suggest that VLM-based relational annotations could complement detector-based supervision by covering higher-order visual semantics. However, this is a limited annotation-level proof-of-concept: the candidate phrases are caption-derived, and GPT-4o is used in both concept construction and visual evaluation, although the evaluation excludes captions.

\subsection{Vocabulary Coverage and Semantic Validity}
\label{app:semantic_limitations}

\paragraph{Fine-grained zero-shot coverage.}
The zero-shot model is trained on CC12M with a fixed, task-agnostic vocabulary of 15,000 concepts, without target-specific vocabulary construction or parameter fitting. On CUB, MM-CBM achieves 36.3\% accuracy compared with 62.3\% for CLIP. This gap may partly reflect limited coverage of discriminative bird attributes: although the model accepts unseen class names as textual queries, distinguishing fine-grained categories requires the relevant attributes to be represented in its concept vocabulary.

\paragraph{Coordinate-level semantic alignment.}
We conduct a blind VLM-based audit of 60 randomly sampled CUB concepts to assess whether learned coordinates align with their assigned concept names. For each concept, GPT-4o compares a group of top-activation images with a randomly selected control group, given only the concept name and images, without activation values, category labels, or group identities. The concept-match rate is 66.67\% for top-activation images and 30.83\% for random controls, and GPT-4o correctly identifies the high-activation group in 76.67\% of cases. These results indicate that the sampled coordinates capture semantic information aligned with their concept names.

\paragraph{Scope and future directions.}
The audit provides evidence of semantic alignment at the image-group level. Extending this evaluation to individual activations and broader datasets would help distinguish responses to the named concept from responses to correlated attributes or contextual cues. This complements intervention and decision-faithfulness evaluations, which assess how concepts influence predictions. Comprehensive single-coordinate semantic auditing remains an open challenge.
}

\section{Additional Experiments}
\subsection{Representation and Hyperparameter Ablations}
\label{Ap: ablation}

\begin{figure}[h]
  \centering
  \includegraphics[width=0.45\linewidth]{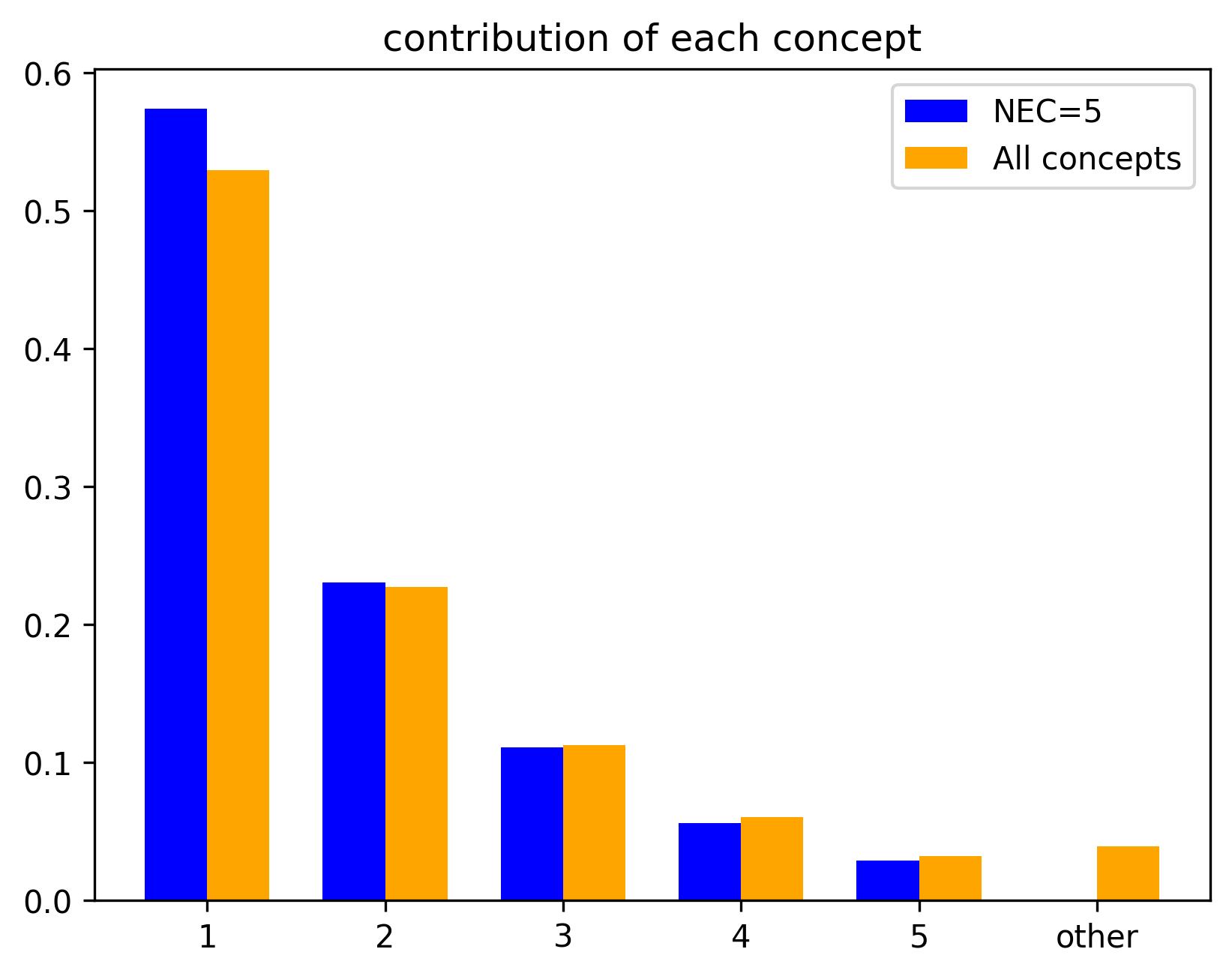}
  \caption{Contribution of each concept used as explanation.}
  \label{fig:ablation nec}
\end{figure}

\noindent\textbf{Ablation study: number of effective concepts.}
We examine the effect of the number of effective concepts (NEC) by comparing performance under $\text{NEC}=5$ with using all concept activations for computing alignment scores. Specifically, we measure the ratio of the top five activations to the total contribution. As noted by~\cite{srivastavavlg}, CBMs trained with sparse concept supervision tend to rely on a small set of high-response concepts, improving robustness to NEC variations. Our results in Fig.~\ref{fig:ablation nec} exhibit the same tendency: even when all activations are used, the top two dominate the decision. Enforcing $\text{NEC}=5$ further sharpens the activation distribution and increases variance, indicating that the model selects more distinct and informative concepts. This property can be leveraged to refine the concept set, producing explanations that are both concise and interpretable.

\begin{figure}[h]
  \centering
  \includegraphics[width=0.35\linewidth]{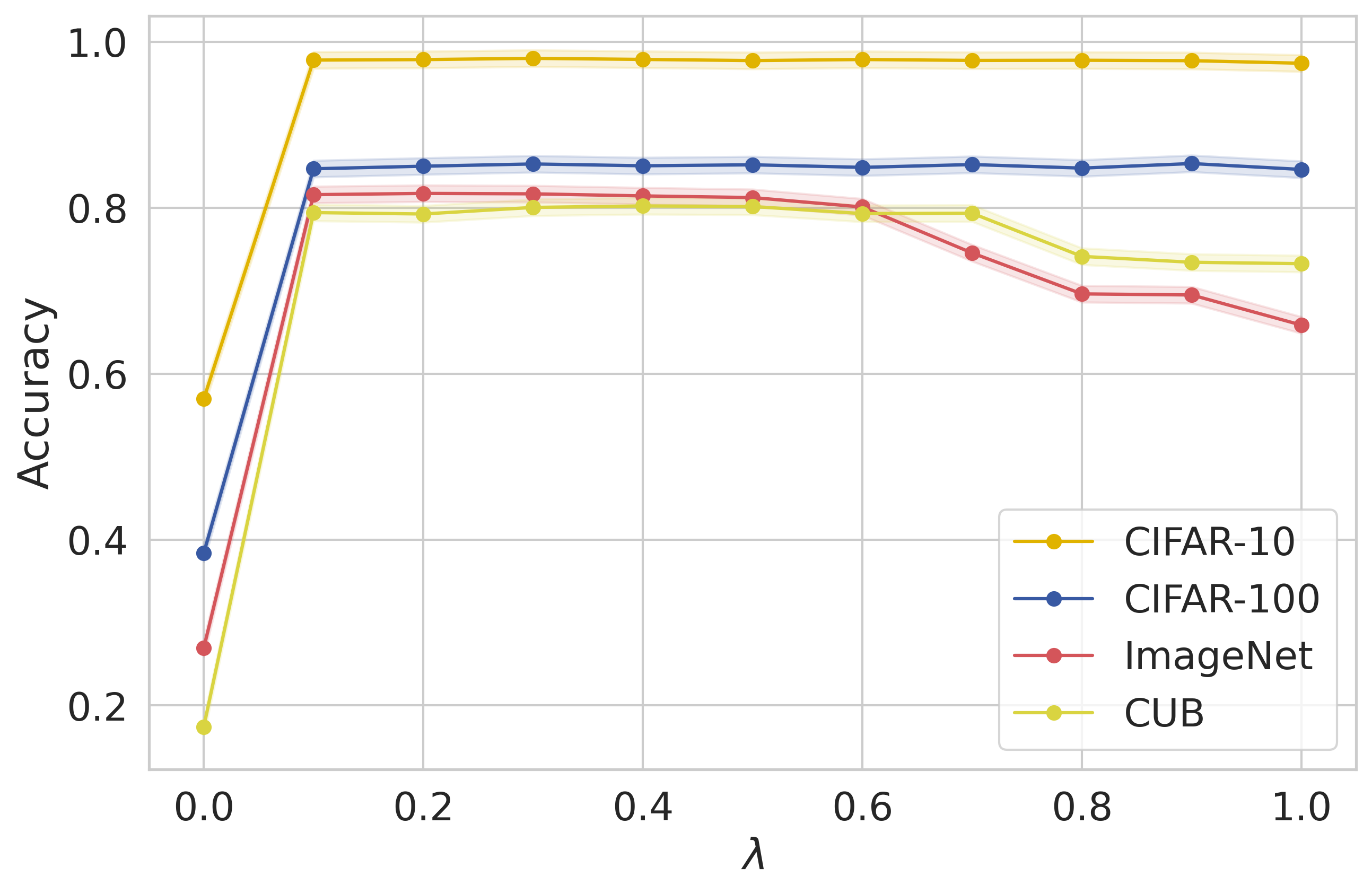}
  \caption{Accuracy across different $\lambda$ value.}
  \label{fig:ablation lambda}
\end{figure}

\noindent\textbf{Ablation study: trade-off weight $\lambda$.}
We further analyze the influence of the trade-off weight $\lambda$ that balances task loss and interpretability loss. As illustrated in Fig.~\ref{fig:ablation lambda}, incorporating the interpretability loss consistently improves performance relative to optimizing the task loss alone. This effect likely arises because concept-based supervision provides richer semantic structure than raw image–label pairs, guiding the model toward human-aligned semantics and improving generalization. Remarkably, even when trained solely with the interpretability loss, MM-CBM still retains non-trivial task performance across four datasets, demonstrating the strong semantic learning capability induced by concept-level alignment.

\begin{figure}[h]
  \centering
  \includegraphics[width=0.35\linewidth]{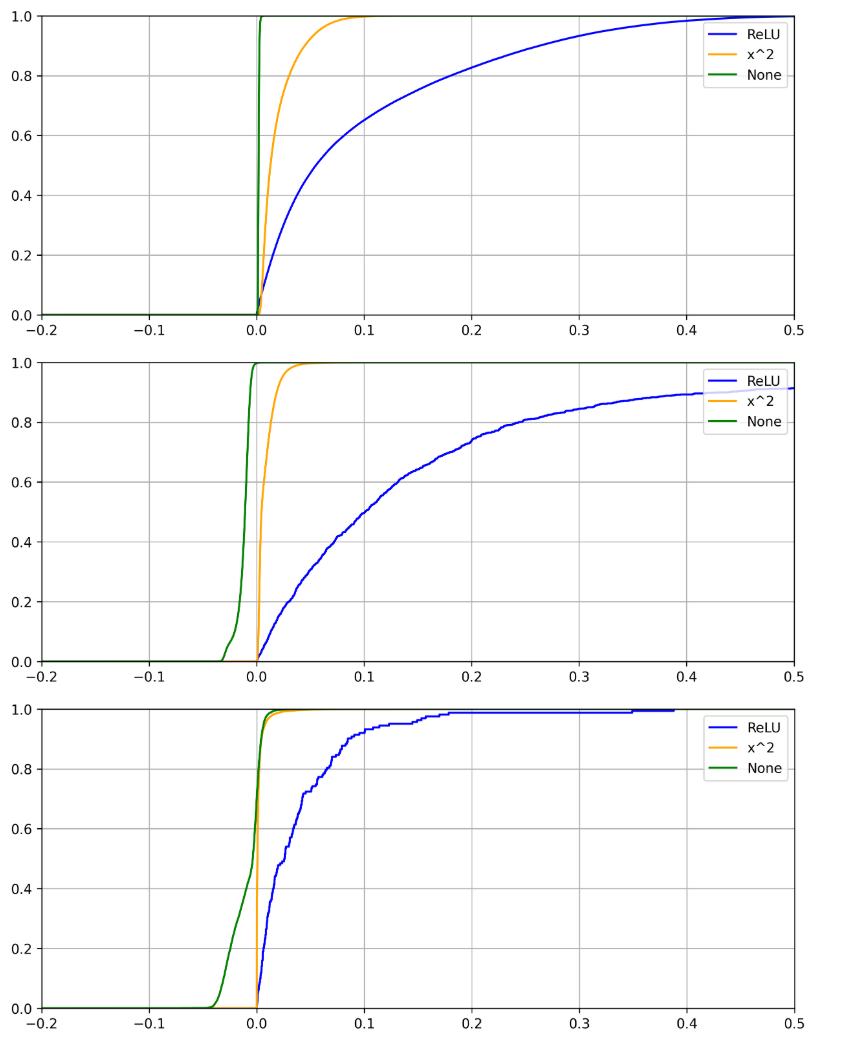}
  \caption{Non-zero concept activation cumulative distribution function of final prediction(top), visual activation(middle) and language activation(bottom).}
  \label{fig:ablation am}
\end{figure}

\noindent\textbf{Ablation study: non-negative concept representation space.}
We evaluate the influence of enforcing non-negative concept activations by comparing alternative transformation functions to the ReLU baseline described in Section~\ref{Ap: enhance}. In particular, we include a squared activation function that guarantees non-negativity. To understand the effect of these transformations, we analyze the cumulative distribution functions (CDFs) of activation values for visual activations, text activations, and final decision activations. As shown in Fig.~\ref{fig:ablation am}, ReLU produces the highest overall activation magnitudes. Text activations, supervised via text similarity, exhibit tighter concentration and lower variance, whereas visual activations, trained via one-hot labels, show higher dispersion, enhancing interpretability by producing clearer concept separation. Moreover, the dot-product operation effectively filters redundant textual information, allowing decision activations to retain higher variance and facilitating the selection of salient, task-relevant concepts.

\clearpage

\subsection{Concept Vocabulary Size and Selection}
\label{app:concept_size}

We investigate how the size of the concept vocabulary affects MM-CBM performance.  A larger concept vocabulary provides more semantic capacity, but may also introduce redundant concepts. Therefore, we evaluate whether performance gains come from increasing vocabulary size or from selecting more informative concepts.

\paragraph{Concept selection strategy.}

For a concept dimension $j$, we define its importance score based on the average positive activation from the text-side Concept Bottleneck Layer:

\begin{equation}
s_j =
\frac{1}{|\mathcal{Y}|}
\sum_{y\in\mathcal{Y}}
\max(0, g_T(e_y)_j),
\end{equation}

where $\mathcal{Y}$ denotes the set of class names, $e_y$ is the CLIP text
embedding of class $y$, and $f_T(\cdot)$ represents the text CBL. Concepts
are ranked according to $s_j$, and the top-$n$ concepts are selected to
construct nested vocabularies:

\begin{equation}
C^{k_1}\subset C^{k_2}\subset ... \subset C^{\mathrm{full}} .
\end{equation}

\paragraph{Evaluation settings.}

We compare three settings:

\begin{itemize}
    \item \textbf{Ranked retraining}: select the top-$n$ concepts according to the importance score and retrain MM-CBM using only these concepts.

    \item \textbf{Random retraining}: randomly sample nested concept subsets with the same vocabulary sizes and retrain MM-CBM. This controls for the effect of vocabulary size alone.

    \item \textbf{Ranked post-hoc pruning}: directly remove unused concept dimensions from the full-vocabulary checkpoint without retraining. This evaluates whether reducing the vocabulary can be achieved by simple
    coordinate pruning.
\end{itemize}

\paragraph{ImageNet}
The full ImageNet vocabulary contains 4,751 concepts. Table~\ref{tab:imagenet_concept_size}
shows the effect of vocabulary size.

\begin{table}[h]
\centering
\caption{
Concept vocabulary size ablation on ImageNet.
Ranked and random settings retrain MM-CBM with reduced vocabularies,
while post-hoc directly prunes the full model.
}
\label{tab:imagenet_concept_size}
\begin{tabular}{c|ccc}
\toprule
\# Concepts &
Ranked Retrain &
Random Retrain &
Ranked Post-hoc
\\
\midrule
500  & 69.33 & 64.69 & 63.93 \\
1000 & 77.54 & 74.61 & 76.56 \\
2000 & 80.38 & 80.01 & 81.75 \\
3000 & 80.89 & 81.51 & 81.76 \\
4751 & 81.74 & -- & 81.74 \\
\bottomrule
\end{tabular}
\end{table}

Increasing the vocabulary size from 500 to 2,000 concepts substantially improves Top-1 accuracy, while further increasing the vocabulary provides only marginal gains. Ranked selection provides the largest improvement under small vocabulary budgets, outperforming random selection by 4.64 and 2.94
percentage points with 500 and 1,000 concepts, respectively. However, this difference diminishes when sufficient concepts are available, suggesting that semantic coverage becomes the dominant factor at larger vocabulary sizes.

\paragraph{CUB}
The full CUB vocabulary contains 370 concepts. Results are shown in Table~\ref{tab:cub_concept_size}.

\begin{table}[h]
\centering
\caption{
Concept vocabulary size ablation on CUB.
}
\label{tab:cub_concept_size}
\begin{tabular}{c|ccc}
\toprule
\# Concepts &
Ranked Retrain &
Random Retrain &
Ranked Post-hoc
\\
\midrule
50  & 38.62 & 10.92 & 10.10 \\
100 & 48.08 & 24.25 & 18.76 \\
200 & 60.16 & 65.54 & 48.62 \\
300 & 70.74 & 72.44 & 68.43 \\
370 & 76.46 & -- & 76.46 \\
\bottomrule
\end{tabular}
\end{table}

Compared with ImageNet, CUB benefits more from increasing the concept vocabulary size due to its fine-grained recognition requirement. Ranked selection provides substantial improvements under limited vocabulary budgets, outperforming random selection by 27.70 and 23.83 percentage points with 50 and 100 concepts. At larger vocabulary sizes, random subsets become
competitive or slightly better, which correlates with their larger concept supervision coverage.

At a fixed vocabulary size, ranked and random subsets have the same bottleneck dimensionality and training protocol, differing only in which concepts are retained. The superior performance of ranked subsets under small concept budgets therefore shows that the gain cannot be attributed solely to vocabulary size. Instead, the semantic composition and supervision coverage of the selected vocabulary materially affect predictive performance. However, this comparison evaluates vocabulary relevance rather than directly establishing that every learned CBL coordinate faithfully represents its assigned concept; we evaluate coordinate-level semantic alignment separately.

\paragraph{Supervision Coverage Analysis}

To further understand the behavior of ranked and random selection on CUB, we
measure the percentage of images containing at least one positive concept
within the selected vocabulary.

\begin{table}[h]
\centering
\caption{
Percentage of CUB images covered by at least one selected positive concept.
}
\label{tab:cub_concept_coverage}
\begin{tabular}{c|cc}
\toprule
\# Concepts & Ranked & Random \\
\midrule
50  & 44.80 & 22.59 \\
100 & 58.91 & 46.56 \\
200 & 82.90 & 85.41 \\
300 & 88.64 & 94.15 \\
\bottomrule
\end{tabular}
\end{table}

These results indicate that importance-based selection is especially useful
when the concept budget is highly constrained. When the vocabulary becomes
larger, random subsets may achieve comparable performance because they provide
higher supervision coverage.

\subsection{Learning with Knowledge Distillation}
\label{Ap: KD}

\begin{table*}[h!]

\centering
\caption {Knowledge distillation accuracy comparison with black-box CLIP.}

\begin{tabular}{lccccccc}
\toprule
    {Method}  & \multicolumn{7}{c} {Dataset}\\
    \cmidrule(lr){2-8}
    {}  & CIFAR-10 & CIFAR-100 & CUB & Food & OxfordPets & DTD & ImageNet \\
    \midrule
    {CLIP ViT-L/14} & 96.2 & 77.9 & 62.3 & 92.9 & 93.5 & 55.3 & 75.3\\
    {\textbf{MM-CBM w/ KD}} & 91.7 & 73.3 &  61.7  &  92.5  & 88.9 & 34.7 & 74.7\\
\bottomrule

\end{tabular}
\label{table: unsupervised_acc}
\end{table*}
When ground-truth class labels are unavailable, we adopt the predictions of the backbone VLM (e.g., CLIP) as soft supervision. This unsupervised learning strategy enhances the flexibility of our framework, enabling the use of large-scale unlabeled images from the target domain together with only the labels of interest, thereby fully exploiting CLIP’s representation capabilities.  

Inspired by prior work on knowledge distillation~\citep{hinton2015distilling, yang2024clip}, we align the output distributions of our model with those of the VLM in both image-to-text and text-to-image directions. Let $M_{ij} = \text{cos}((I_e)_i, (T_e)_j)$ denote the similarity matrix in the concept space, and $N_{ij} = \cos(E_I(x_i), E_T(y_j))$ the similarity matrix from CLIP. All embeddings are $L_2$-normalized. The corresponding softmax-normalized distributions are:  

\begin{equation}
    p_T = \text{softmax}(N)\text{,}~ p_S = \text{softmax}(N^\top)
\end{equation}

\begin{equation}
    q_T = \text{softmax}(M)\text{,}~ q_S = \text{softmax}(M^\top)
\end{equation}

We then minimize the Kullback–Leibler (KL) divergence between the teacher (CLIP) and student (CBL) distributions:

\begin{align}
    L_{\text{KD}_{\text{I}\rightarrow \text{T}}} &= D_{\text{KL}}(p_T \| q_T) = \sum_i p_T(i) \log \frac{p_T(i)}{q_T(i)}, \\
    L_{\text{KD}_{\text{T}\rightarrow \text{I}}} &= D_{\text{KL}}(p_S \| q_S) = \sum_i p_S(i) \log \frac{p_S(i)}{q_S(i)}, \\
    L_{\text{KD}} &= \frac{1}{2} \left( L_{\text{KD}_{\text{I}\rightarrow \text{T}}} + L_{\text{KD}_{\text{T}\rightarrow \text{I}}} \right).
\end{align}

Additionally, we treat CLIP’s top-1 prediction as a pseudo-label $\hat{l}^{\mathrm{CLIP}}$ to supervise the classification head: 
\begin{equation}
    L_{\text{ACC}}^{\text{KD}} = 
    L_{\text{CE}}\left( \frac{I_e \cdot T_e}{\|I_e\|_2 \|T_e\|_2} \cdot e^{\tau},\ \hat{l}^{\mathrm{CLIP}} \right) 
    + L_{\text{KD}}.
\end{equation}

This unsupervised task-performance loss can be directly incorporated into the final objective in Equation~\ref{Eq: final_loss}, replacing the supervised loss, thereby enabling end-to-end training of an interpretable CLIP without requiring labeled data.  

As shown in Table~\ref{table: unsupervised_acc}, the knowledge-distilled MM-CBM largely preserves task performance across the other six datasets. In contrast, its performance on the DTD dataset is noticeably weaker. A plausible explanation is that the black-box model itself performs poorly on DTD, resulting in soft labels that lack sufficiently informative latent knowledge, which in turn limits the effectiveness of the distilled model. This result demonstrates the strong scalability of our approach: given an image and an associated category of interest, it can achieve performance close to that of the black-box model, thereby greatly broadening the range of potential applications for MM-CBM.

\subsection{Evaluation with Alternative Backbones}
\label{Ap: backbone}
To evaluate the generalization capability of our approach, we conduct experiments with multiple variants of CLIP as well as other backbone models such as DINOv2. Since DINOv2 only provides an image encoder, we incorporate a SigLIP2 text encoder to facilitate the training process. The results in Table~\ref{table: other backbone} demonstrate that our method can be seamlessly adapted to different architectures with similar designs, rather than being restricted to interpretable variants of CLIP.
\begin{table}[h]
\centering
\caption {Accuracy comparison with stronger baseline under finetuned settings.}
\scalebox{1}{
\begin{tabular}{lccccccc}
\toprule
    {Method} & \multicolumn{7}{c} {Dataset}\\
    \cmidrule(lr){2-8}
    {Accuracy} & CIFAR10 & CIFAR100 & CUB & DTD & Food & Pets & ImageNet \\
    \midrule
    {EVA-CLIP ViT-L/14} & 99.5 & 93.0 & 84.3 & 81.1 & 94.8 & 94.4 & 85.3 \\
    {\textbf{MM-CBM}} & 99.3 & 92.7 & 82.8 & 82.0 & 94.3 & 93.6 & 83.8 \\
    \midrule
    {SigLIP ViT-L/16} & 97.8 & 87.4 & 84.4 & 82.8 & 95.4 & 94.9 & 86.4 \\
    {\textbf{MM-CBM}} & 98.0 & 87.2 & 79.9 & 82.1 & 95.6 & 93.7 & 85.1 \\
    \midrule
    {SigLIP2 ViT-L/16} & 98.1 & 86.9 & 83.7 & 79.7 & 95.8 & 96.2 & 87.1 \\
    {\textbf{MM-CBM}} & 98.2 & 86.4 & 77.8 & 82.8 & 95.9& 93.9 & 85.7 \\
    \midrule
    {DINOv2 ViT-L/14} & 99.1 & 91.0 & 86.7 & 78.3 & 92.7 & 96.0 & 84.5 \\
    {\textbf{MM-CBM}} & 99.2 & 91.2 & 84.6 & 78.0 & 93.0 &  94.5 & 83.1 \\

\bottomrule
\end{tabular}
}
\label{table: other backbone}
\end{table}




\clearpage

\section{Concept-Based Explanations and Interventions}
\label{rebuttal}
\subsection{Decision Faithfulness}
\label{app:faithfulness}

To evaluate whether the highlighted concepts faithfully reflect the model's decision process, we perform sufficiency and comprehensiveness evaluations. Unlike preference-based human or VLM evaluations, these metrics directly measure whether the selected concepts are sufficient to preserve predictions and whether they are necessary for maintaining the model's decisions.

For each sample, concept contributions are computed based on the contribution of each concept dimension to the final image-text similarity score. Concepts are ranked according to their contribution magnitude.

\paragraph{Sufficiency evaluation.}

Given the top-$n$ contributing concepts, we retain only these concepts and remove all remaining concept dimensions before classification. A faithful explanation should preserve the prediction when the retained concepts contain the main evidence used by the model.

\paragraph{Comprehensiveness evaluation.}

We remove the top-$n$ contributing concepts and evaluate the resulting prediction accuracy. A faithful explanation should cause a larger performance degradation when the most important concepts are removed compared with random concept removal.

For both evaluations, we compare against random concept selections with the same number of concepts with positive contribution to the prediction. Experiments are performed on the full ImageNet validation set (50K images) and the full CUB test set (5,790 images). 

\paragraph{Effect of sparse concept usage.}

It is worth noting that the random baselines in these experiments are relatively strong controls because MM-CBM performs prediction through a sparse set of active concepts. In particular, each image-text prediction typically depends only on a small number of concepts (approximately ten candidate concepts before selecting the top contributing ones). Therefore, randomly retaining or removing concepts has a non-negligible probability of overlapping with decision-relevant concepts.

Despite this challenging setting, the top contributing concepts consistently outperform random selections in both retain and remove evaluations. This behavior further supports the sparse nature of MM-CBM predictions: a small subset of concepts captures most of the decision evidence, while the remaining concept dimensions contribute less to the final prediction.

\paragraph{ImageNet}

The ImageNet model achieves 81.74\% Top-1 accuracy without intervention. Table~\ref{tab:faithfulness_imagenet} reports the sufficiency and comprehensiveness results. The results show that the top contributing concepts contain substantially more decision-relevant information than random concepts. For example, retaining the top-3 concepts preserves 79.16\% Top-1 accuracy, compared with 54.91\% for random concepts. Conversely, removing the top-3 concepts reduces accuracy to 12.87\%, while removing random concepts only reduces accuracy to 47.80\%. 

\begin{table}[h]
\centering
\caption{
Sufficiency and comprehensiveness evaluation on ImageNet validation set. Top concepts are selected according to model contribution scores, while random controls use randomly selected positive contributed concepts with matched cardinality.
}
\label{tab:faithfulness_imagenet}
\begin{tabular}{c|cc|cc}
\toprule
 & \multicolumn{2}{c|}{Sufficiency (Top-1 $\uparrow$)}
 & \multicolumn{2}{c}{Comprehensiveness (Top-1 $\downarrow$)}
\\
$k$ &
Top retain &
Random retain &
Top remove &
Random remove
\\
\midrule
1 & 47.44 & 18.34 & 59.91 & 74.51 \\
2 & 70.73 & 36.15 & 31.62 & 62.48 \\
3 & 79.16 & 54.91 & 12.87 & 47.80 \\
5 & 81.72 & 76.42 & 3.86 & 22.08 \\
\bottomrule
\end{tabular}
\end{table}

\paragraph{CUB}

The CUB model achieves 76.46\% Top-1 accuracy without concept removal.
Table~\ref{tab:faithfulness_cub} reports the corresponding results. Similar trends are observed on CUB. The top contributing concepts consistently provide higher sufficiency and stronger comprehensiveness than random controls. For example, retaining the top-3 concepts preserves 72.59\% Top-1 accuracy compared with 57.24\% for random concepts. Removing these concepts decreases accuracy to 7.58\%, compared with 27.24\% under random removal. When retaining the top-5 concepts, the model recovers the full prediction accuracy (76.46\%), suggesting that the selected concepts capture most of the decision evidence used by the model.

\begin{table}[h]
\centering
\caption{
Sufficiency and comprehensiveness evaluation on the CUB test set.
}
\label{tab:faithfulness_cub}
\begin{tabular}{c|cc|cc}
\toprule
 & \multicolumn{2}{c|}{Sufficiency (Top-1 $\uparrow$)}
 & \multicolumn{2}{c}{Comprehensiveness (Top-1 $\downarrow$)}
\\
$k$ &
Top retain &
Random retain &
Top remove &
Random remove
\\
\midrule
1 & 40.36 & 19.21 & 43.16 & 58.79 \\
2 & 62.28 & 38.91 & 20.90 & 41.19 \\
3 & 72.59 & 57.24 & 7.58 & 27.24 \\
5 & 76.46 & 75.04 & 2.02 & 8.51 \\
\bottomrule
\end{tabular}
\end{table}

\subsection{Concept Interventions}
\label{app:intervention}

We evaluate three intervention operators on the visual concept representation:

\begin{itemize}
    \item \textbf{Binary replacement:} replace the complete visual CBL representation with the binary positive concept-label vector.
    \item \textbf{Suppress unlabeled:} retain the original activations only for positively labeled concepts and set all other coordinates to zero.
    \item \textbf{Positive addition:} add $s\delta$ to each positively labeled concept coordinate, where $s \in \{0.5,1,2\}$ and $\delta$ is the 90th percentile of positive visual CBL activations.
\end{itemize}

For all interventions, we re-normalize the modified concept representation and rerun the full NEC-$5$ classifier. 

\begin{table}[h]
\centering
\caption{Systematic concept intervention results on the full ImageNet validation set. ``Error correction'' is the fraction of originally incorrect predictions that become correct, while ``collateral damage'' is the fraction of originally correct predictions that become incorrect.}
\label{tab:imagenet_intervention}
\begin{tabular}{lcccc}
\toprule
Intervention & Top-1 (\%) & Error correction (\%) & Collateral damage (\%) & Net correction \\
\midrule
Baseline & 81.738 & -- & -- & -- \\
Binary replacement & 49.414 & 45.71 & 49.76 & -16,162 \\
Suppress unlabeled & 50.726 & 26.96 & 43.96 & -15,506 \\
Positive add ($0.5\delta$) & \textbf{87.058} & 41.08 & \textbf{2.67} & \textbf{+2,660} \\
Positive add ($1.0\delta$) & 84.686 & 51.06 & 7.80 & +1,474 \\
Positive add ($2.0\delta$) & 77.258 & 52.81 & 17.28 & -2,240 \\
\bottomrule
\end{tabular}
\end{table}

The baseline Top-5 accuracy is 94.872\%, which increases to 97.278\% under the best intervention. Here, $\delta=6.5971$.

\begin{table}[h]
\centering
\caption{Systematic concept intervention results on the full CUB test set.}
\label{tab:cub_intervention}
\begin{tabular}{lcccc}
\toprule
Intervention & Top-1 (\%) & Error correction (\%) & Collateral damage (\%) & Net correction \\
\midrule
Baseline & 76.459 & -- & -- & -- \\
Binary replacement & 58.636 & 54.44 & 40.07 & -1,032 \\
Suppress unlabeled & 52.504 & 12.03 & 35.04 & -1,387 \\
Positive add ($0.5\delta$) & \textbf{83.161} & 60.67 & \textbf{9.92} & \textbf{+388} \\
Positive add ($1.0\delta$) & 77.219 & 61.12 & 17.82 & +44 \\
Positive add ($2.0\delta$) & 71.744 & 59.43 & 24.46 & -273 \\
\bottomrule
\end{tabular}
\end{table}

The baseline Top-5 accuracy is 90.121\%, which increases to 94.335\% under the best intervention. Here, $\delta=2.2413$.

The poor performance of binary replacement does not necessarily imply that the learned concept coordinates lack semantic meaning. Binary replacement is a stringent intervention that discards the continuous CBL representation and replaces it with a sparse binary annotation vector. This operation removes activation magnitudes, introduces a substantial distribution shift, and treats undetected concepts as absent despite the incomplete recall of automatically constructed annotations. The results therefore indicate that the binary concept labels are insufficient as a complete replacement for the learned representation. In contrast, mild positive interventions preserve the original representation while injecting semantically aligned evidence, yielding a more favorable correction--collateral-damage trade-off.

\paragraph{Semantic permutation control.}
We further permute the concept coordinates using five random seeds while preserving the number of positive labels per image and the overall concept-frequency distribution. This control breaks the semantic correspondence between concept labels and CBL coordinates. Under the $0.5\delta$ intervention, the random-control Top-1 accuracy is 81.080\% on ImageNet and 53.537\% on CUB, compared with 87.058\% and 83.161\% under the semantically aligned intervention. The corresponding improvements over the random controls are 5.978 and 29.623 percentage points, with 95\% confidence intervals of $[5.699, 6.242]$ and $[28.480, 30.795]$, respectively.

\paragraph{Qualitative expert-guided interventions.}

\begin{figure}[h]
  \centering
  \includegraphics[width=\linewidth]{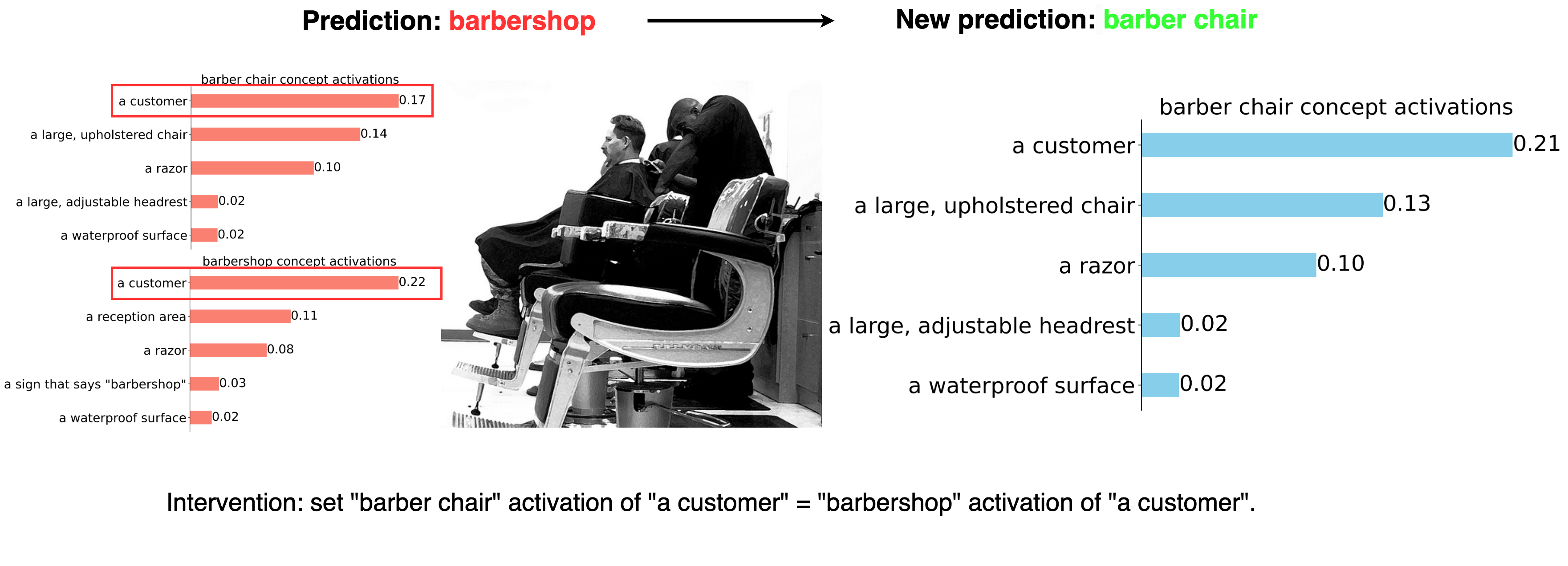}
  \caption{A sample of correcting model prediction by setting the common concept to the same value}
  \label{fig:intervention_sup}
\end{figure}

Another error occurs when ``barbershop'' is incorrectly favored due to a higher response to ``a customer,'' despite ``barber chair'' being the correct label. Equalizing the activation between both classes for that concept ($T_{e[\text{gt, concept}]} = T_{e[\text{pred, concept}]}$) corrects 5 predictions and introduces 2 new errors, shifting predictions from ``barbershop'' to ``barber chair.'' This leads to a 3\% accuracy improvement in a 100-sample subset. Such errors stem from \textbf{response bias on shared concepts}, which hinders fine-grained classification when dominant but insufficient features overshadow more specific ones. The source of this error is traceable: since $I_e$ is identical, the difference lies in $T_e$. Using the label generation model \texttt{all-mpnet-base-v2}~\citep{wolf2020transformers}, we find that the similarity score between ``barbershop'' and ``a customer'' is 0.3618, compared to 0.3070 for ``barber chair.'' This discrepancy reflects a language-model-induced bias during training.

\subsection{Qualitative Image Retrieval Results}
\label{Ap: ir_sup}
\begin{figure*}[h]
  \centering
  \includegraphics[width=\linewidth]{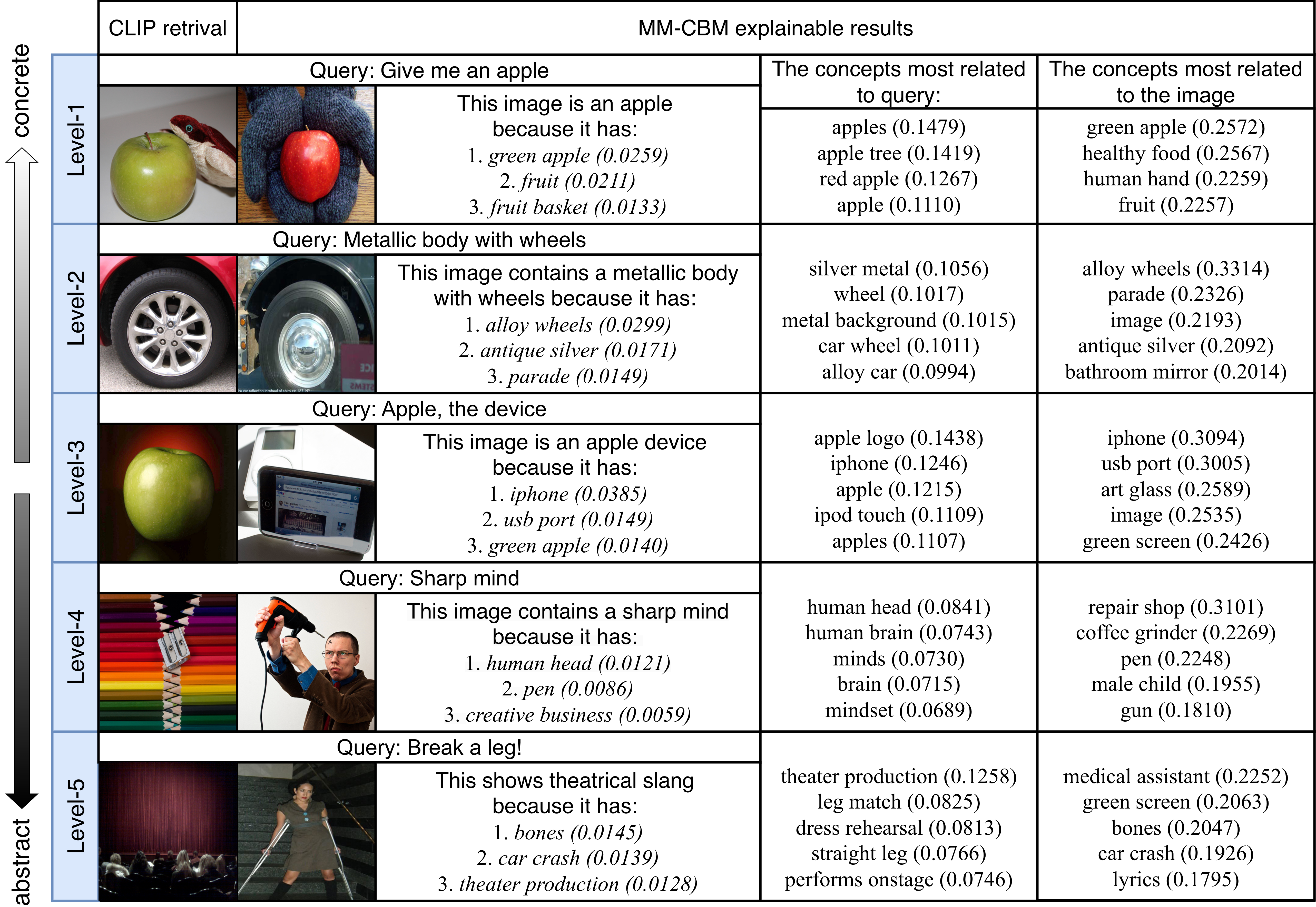}
  \caption{Image Retrieval on five different types of queries (failure cases in Level-4 and Level-5).}
  \label{fig:imageretrieval_failurecase}
\end{figure*}

\begin{figure*}[h]
  \centering
  \includegraphics[width=\linewidth]{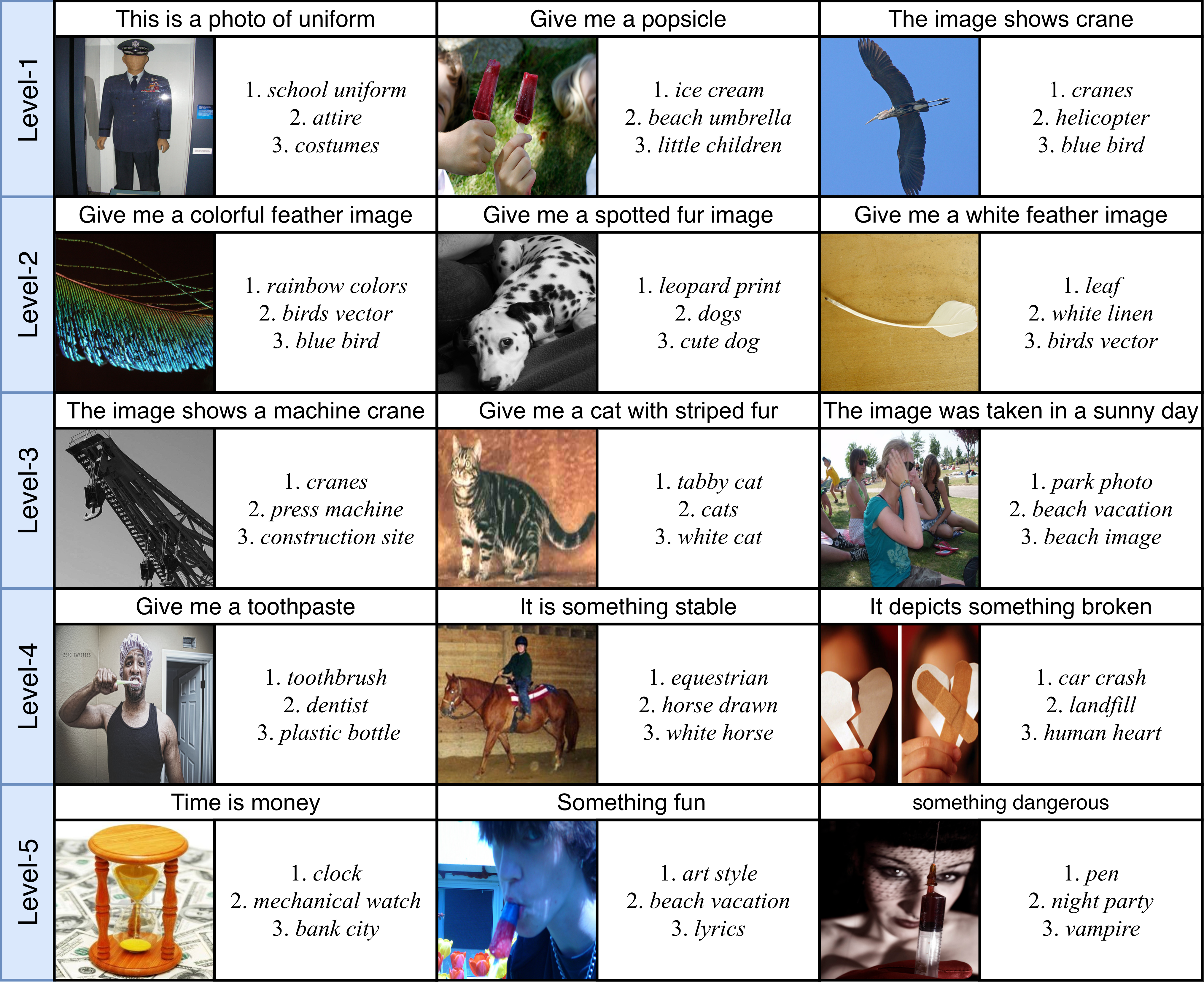}
  \caption{Image Retrieval on five different types of queries. The top of the module shows the query statement we use, and the right side shows the most relevant concepts.}
  \label{fig:imageretrieval_sup}
\end{figure*}

In Section~\ref{sec:usecase}, we define 5 different levels of queries and provide corresponding examples. In this section, we provide more cases and offer interpretable predictions in Fig.~\ref{fig:imageretrieval_sup} and Fig.~\ref{fig:imageretrieval_failurecase}.

\stopcontents[appendices]

\end{document}